\documentclass{article}

\usepackage{microtype}
\usepackage{graphicx}
\usepackage{subfigure}
\usepackage{booktabs} 

\usepackage{hyperref}

\usepackage[accepted]{icml2021}

\usepackage{amssymb}
\usepackage{amsmath}
\usepackage{comment}
\usepackage{caption}

\usepackage{bbm}
\usepackage{mathtools}
\usepackage{dblfloatfix}  
\DeclareMathOperator*{\argmax}{\arg\!\max}

\DeclarePairedDelimiter\ceil{\lceil}{\rceil}
\DeclarePairedDelimiter\floor{\lfloor}{\rfloor}
\newcommand{\tvp}{\text{TV}_{p}}
\newcommand{\tvw}{\text{TV}_{w}}
\newcommand{\klw}{ \text{KL}_{w}}
\newcommand{\klp}{\text{KL}_{p}}

\icmltitlerunning{Statistical Distance Based Deterministic Offspring Selection in SMC Methods}

\begin{document}

\twocolumn[
\icmltitle{Statistical Distance Based Deterministic Offspring Selection in SMC Methods}



\icmlsetsymbol{equal}{*}

\begin{icmlauthorlist}
\icmlauthor{Oskar Kviman}{to,goo}
\icmlauthor{Hazal Koptagel}{to,goo}
\icmlauthor{Harald Melin}{to,goo} 
\icmlauthor{Jens Lagergren}{to,goo}
\end{icmlauthorlist}

\icmlaffiliation{to}{Department of Electrical Engineering and Computer Science, KTH Royal Institute of Technology, Stockholm, Sweden}
\icmlaffiliation{goo}{Science for Life Laboratory, Stockholm, Sweden}

\icmlcorrespondingauthor{Oskar Kviman}{okviman@kth.se}
\icmlcorrespondingauthor{Jens Lagergren}{jens.lagergren@scilifelab.se}

\icmlkeywords{Machine Learning, ICML, SMC, pMCMC, Resampling}

\vskip 0.3in
]



\printAffiliationsAndNotice{}  

\begin{abstract}
Over the years, sequential Monte Carlo (SMC) and, equivalently, particle filter (PF) theory has gained substantial attention from researchers. However, the performance of the resampling methodology, also known as offspring selection, has not advanced recently. We propose two deterministic offspring selection methods, which strive to minimize the Kullback-Leibler (KL) divergence and the total variation (TV) distance, respectively, between the particle distribution prior and subsequent to the offspring selection. By reducing the statistical distance between the selected offspring and the joint distribution, we obtain a heuristic search procedure that performs superior to a maximum likelihood search in precisely those contexts where the latter performs better than an SMC.
For SMC and particle Markov chain Monte Carlo (pMCMC), our proposed offspring selection methods always outperform or compare favorably with the two state-of-the-art resampling schemes on two models commonly used as benchmarks from the literature.
\end{abstract}

\section{Introduction}

Since the bootstrap particle filter (BPF) and multinomial resampling were both proposed in the seminal work of \citet{gordon1993novel}, sequential Monte Carlo (SMC) methods have been heavily researched and applied to a variety of problems, such as robotics \cite{grisettiyz2005improving, eade2006scalable}, phylogenetic tree inference \cite{bouchard2012phylogenetic}, econometrics \cite{aux_PF}, and much more. There are two important explanations for its popularity. First, given a Bayesian model of any complex, non-linear or/and non-Gaussian dynamical system, SMC algorithms, interchangeably referred to as particle filters (PFs), produce impressive solutions without simplification of  the model. Second, it is straightforward to understand, at least informally, how the PF works.

Following the invention of the BPF, advanced improvements have been proposed. The BPF is likely the simplest possible PF as it employs the model prior as its proposal distribution, and so, in attempts to obtain more sophisticated PFs, designing clever proposals has gained attention. For instance, the auxiliary PF \cite{aux_PF} utilizes future observations in order to improve the proposals. Not long ago, \citet{naesseth2017variational} presented a proposal distribution with parameters which were tuned using variational inference (VI). Another example of combining VI and SMC is given in \citet{fivo}, where they derive an alternative to the evidence lower bound (ELBO) by letting a variant of the PF's log-likelihood be an objective function, which can be used for gradient descent minimization. Also, \citet{saeedi2017variational} attempt to simultaneously update a complete set of trajectories from an earlier PF run by maximizing the ELBO objective. But as their method takes a posterior approximation as input, it is more appropriately compared with, e.g., a particle Markov chain Monte Carlo (pMCMC) method. Although they do not utilize VI, both \citet{fox2002kld} and \citet{soto2005self} use the Kullback-Leibler (KL) divergence between the target distribution (making simplifying assumptions on the form of the target distribution) and the approximation to adapt the number of particles in the PF.

However, it appears that since VI's rise in popularity in the late 90's \cite{jordan1999introduction}, little attention has been devoted to SMC's offspring selection – often referred to as the resampling step, as it is typically done stochastically.
Common resampling schemes, all stochastic, are; stratified and systematic resampling \cite{kitagawa96}, multinomial resampling \cite{gordon1993novel} and residual resampling \cite{higuchi1997monte, liu1998sequential}. In this work, we will use the stratified and systematic resampling schemes as baselines, since they are seemingly the most popular \cite{douc2005comparison,doucet2001introduction,pmcmc}. Deterministic resampling schemes exist, albeit they do often not compare with the standard stochastic ones (mentioned above) in terms of estimation capabilities  \cite{kitagawa96}, or they are not off-the-shelf methods and are difficult to fit into the established theoretical framework surrounding SMC methods \cite{li2012deterministic}.

In this work, we present a set of novel methods\footnote{Code available at \href{https://github.com/Lagergren-Lab/KL-TV-Reshuffling}{github.com/Lagergren-Lab/KL-TV-Reshuffling}} for performing offspring selection in PFs, based on minimization  of  statistical distances. Specifically, we design (1) a reshuffling algorithm that given a normalized particle distribution finds an unweighted particle distribution with minimum KL divergence and (2) another reshuffling algorithm that given the same input finds an unweighted particle distribution with minimal total variation (TV) distance. We prove in the Supplementary Materials that each of  these two reshuffling algorithms attains unweighted particle distributions on minimal statistical distance. Our reshuffling schemes are off-the-shelf methods, ready to be used in place of standard resampling techniques – without manipulation of the SMC algorithm.

\section{Notation and Background Information}
In this section, we introduce the BPF and the state-of-the-art offspring selection schemes.

\subsection{The Bootstrap Particle Filter}
The PF approximates the intractable posterior distribution, $p(x_{1:n} | y_{1:n})$, via a particle distribution \cite{doucet2009tutorial, doucet2001introduction}
\begin{equation}
\label{eq:particle_dist}
    q(x_{1:n}) = \sum_{s} w_n^s \delta^s(x_{1:n}),
\end{equation}
where $\delta^s$ denotes a Dirac distribution at point $s \in S$, and $w_n^s$ is the $s$'th particle's normalized importance weight at time $n \in N$. The unnormalized weight is 

\begin{equation}
\label{eq:weight_updates}
\begin{split}
    \tilde w_n &= \frac{p(x_{1:n},y_{1:n})}{\pi(x_{1:n})} = \frac{g(y_n|x_n)f(x_n|x_{n-1})}{\pi(x_{n}|x_{1:n-1})}
    \tilde w_{n-1}^{1 - \mathbbm{1}_{R_n}},
\end{split}
\end{equation}

where $g(y_n|x_n)$ and $f(x_n|x_{n-1})$ are the model's emission and transition probabilities, respectively, $\mathbbm{1}_{R_n}$ is the indicator function, and $R_n$ is the event that resampling occurred at time $n$\footnote{Setting the $\tilde w_{n-1}$ to 1 or $1/S$ will not matter as $\tilde w_n$ is later normalized.}.

Throughout this work, we will only consider the transition distribution as our proposal distribution, i.e. $\pi(x_n|x_{n-1}) = f(x_n|x_{n-1})$. Doing so, we end up with BPF, the simplest possible PF, and Equation \ref{eq:weight_updates} now becomes

\begin{equation}
\label{eq:weight_adaptive}
    \tilde w_n = g(y_n|x_n)
    \tilde w_{n-1}^{1 - \mathbbm{1}_{R_n}}.
\end{equation}

Although more advanced PFs exists, the BPF's simplicity makes it useful for demonstrating novel techniques – i.e. if a new resampling/reshuffling scheme can boost the performance of the BPF, it is reasonable that more advanced methods should benefit it too. 

An algorithmic description of the BPF is given in Algorithm \ref{alg:bpf}. Note that $i^s_n$ denotes the ancestor index of particle $x^s_{1:n}$, and that $\mathcal{R}$ is an arbitrary \textit{resampling} or \textit{reshuffling} scheme, i.e., an offspring selection method.

\begin{algorithm}[tb]
   \caption{The Bootstrap Particle Filter}
   \label{alg:bpf}
\begin{algorithmic}
   \STATE {\bfseries Input:} $S$, $\{y_n\}_{n=1}^N$
   \STATE {\bfseries Output:} $q(x_{1:N})$
   \STATE initialize $x_1^s \sim\mu(x_1)$
   \STATE compute $\tilde w_1^s = g(y_1|x_1^s)$
   \STATE normalize $w_1^s = \tilde w_1^s / (\sum_s \tilde w_1^s)$
   
   \FOR{$n=2,\ldots,N$}
   \IF{$\sum_s (w^s_{n-1})^{-2} < S / 2$}
   \STATE select ancestors $\{i^s_n\}_{s=1}^S= \mathcal{R}(\{w^s_n\}_{s=1}^S)$
   \STATE set $\tilde w^s_{n-1} = 1$, $\forall s$
   \ELSE 
   \STATE set $i^s_{n} = s$, $\forall s$
   \ENDIF
   \STATE propagate $x^s_n \sim f(x^s_n|x_{n-1}^{i^s_n})$
   \STATE compute $\tilde{w}_n^s = g(y_n|x_n^s) \tilde w_{n-1}^s$ 
   \STATE normalize $w_n^s = \tilde w_n^s / (\sum_s \tilde w_n^s)$
   \ENDFOR
   \STATE \textbf{return} $q(x_{1:N})$
\end{algorithmic}
\end{algorithm} 

\subsection{The Bootstrap Particle Filter with Likelihood}
\label{sec:bpf_likelihood}

A subset of our proposed reshuffling methods works with the particle likelihoods, $p(x_{1:n}, y_{1:n})$. These methods require some updates to the standard BPF algorithm shown in Algorithm \ref{alg:bpf}. To this end, we provide an updated BPF algorithm, \emph{BPF with likelihood}. The algorithm is available in the Supplementary Materials.

\subsection{Offspring Selection}
\label{sec:offspring}

Offspring selection is useful for discarding particles with low importance weights and replacing them with more promising ones. Once a particle $x_{1:n}$ is discarded, a surviving particle is multiplied. This ``circle of life''-like heuristic ensures that there are constantly $S$ particles in the system. Specifically, offspring selection attempts to replace the particle distribution in Equation \ref{eq:particle_dist}, by a distribution in terms of particle multiplicities, 

\begin{equation}
\label{eq:multpl_distr}
    \ddot q(x_{1:n}) = \frac{1}{S} \sum_{s} a_n^s \delta^s(x_{1:n}),
\end{equation}
where $a_n^s\in \mathbb{Z}^+_0$ is the multiplicity of particle $s$ at time $n$. In other words, a traditional resampling scheme attempts to approximate $q(x_{1:n})$ using the set of multiplicities

\begin{equation}
\label{eq:multiplicity_set}
    \mathcal{A}^S = \{a\in(\mathbb{Z}_0^+)^S: \sum_s a^s = S\}.
\end{equation}

The notion of approximating $q(x_{1:n})$, which in turn approximates the target posterior distribution $p(x_{1:n}|y_{1:n})$ might seem confusing, but in order to perform offspring selection, we require a mapping from importance weights to particle multiplicities, $(\mathbb{R}^+_0)^S \mapsto \mathcal{A}^s$.

After resampling, the $S$ offspring are regarded as equiprobable, yielding
\begin{equation}
\label{eq:equi_distr}
    \check q(x_{1:n}) = \frac{1}{S} \sum_{s} \delta^s(x_{1:n}).
\end{equation}
Obviously, $q$ from Equation \ref{eq:particle_dist} is the closest to the target distribution as $\ddot q$ and $\check q$ are mere approximations of $q$. 

In this work, we compare our proposed methods with the state-of-the-art \textit{stratified} and \textit{systematic} resampling schemes. 

\textbf{The stratified resampling} method is inspired by ideas within survey sampling \cite{douc2005comparison}, i.e. samples should also represent areas with less (probability) density, in an attempt to avoid all particles being centered around modes. To perform stratified resampling, one generates $S$ random numbers according to $u_k = ((k -1 ) + \tilde u_k)/S, \quad \tilde u_k \sim \mathcal{U}\{0,1\}.$

These numbers are then used to sample the new index, $s'\in [1, S]$ for each particle, such that $u_{s'} \in \left[
    \sum_{s=1}^{s'-1}w^s_n, \sum_{s=1}^{s'}w^s_n
    \right)$. 

\textbf{Systematic resampling} is similar to the stratified counterpart, only here a single random number is drawn $\tilde u_1 \sim \mathcal{U}\{0, \frac{1}{S}\}$
followed by deterministic assignments of the remaining $S-1$ values, $u_k = \tilde u_1 + (k-1)/S$.

Finally, in order to have a baseline for our likelihood based methods, we introduce a simple maximum likelihood resampling approach, which we will refer to as \textbf{ML}. In ML, only the particle $s$ with the highest likelihood, $p(x_{1:n}^s, y_{1:n})$, is multiplied. 

\section{Kullback-Leibler Reshuffling}
\label{sec:kl_reshuffling}

The KL divergence between two distributions $q$ and $r$, defined as

\begin{equation} 
\label{eq:KL}
    \text{KL}(q(x)||r(x)) = \mathbb{E}_{q}\left[
    \log\frac{q(x)}{r(x)}
    \right],
\end{equation}
measures the similarity of two distributions, is non-negative and only zero when the distributions are identical. 

\subsection{Weight Based KL Reshuffling}
Recall from Section \ref{sec:offspring} that $\ddot q(x_{1:n})$ is an approximation of the $q(x_{1:n})$. The goal of weight based KL reshuffling, $\klw$ is to minimize the KL divergence from $\ddot q(x_{1:n})$ to $q(x_{1:n})$, or equivalently maximize the negative KL divergence. Equation \ref{eq:elbo_klw} shows the function to be maximized. 

\begin{equation}
\label{eq:elbo_klw}
\begin{split}
    \mathcal{L}(x_{1:n}) &= -\text{KL}(\ddot q(x_{1:n})||q(x_{1:n}))\\ &= \sum_s \ddot q(x^s_{1:n}) \log\frac{q(x^s_{1:n})}{\ddot q(x^s_{1:n})}\\
    &\propto \sum_s a^s \log\frac{w^s}{a^s},
\end{split}
\end{equation}

where $a^s$ is the multiplicity of the $s$th particle. 

Algorithm \ref{alg:kl_reshuffling} is our novel algorithm for finding the optimal unweighted particle distribution $\ddot q^*$ the w.r.t. KL divergence. For brevity we let $f(a,s) = a^s \log\frac{u^s}{a^s}$. When the inputs are the importance weights of the particles, $u^s = w^s, \forall s$, the method is referred to as $\klw$. The proof showing that we indeed find $\ddot q^*$ is given in the Supplementary Materials.
 
 \begin{algorithm}[t]
   \caption{Kullback-Leibler Reshuffling}
   \label{alg:kl_reshuffling}
\begin{algorithmic}
   \STATE {\bfseries Input:} $\{u^s\}_{s=1}^S$
   \STATE {\bfseries Output:} $\{i^s\}_{s=1}^S$
   \STATE order $u^1 \geq \ldots \geq u^S$
   \STATE set $a^s = 0$, $\forall s$
   \STATE define $C^+(a, s) = f(a + 1, s) - f(a, s)$
   \WHILE{$\sum_s a^s<S$}
   \STATE compute $t = \argmax_s C^{+}(a_s, s)$ 
   \STATE set $a^t = a^t + 1$
   \ENDWHILE
   \STATE map $\{i^s\}_{s=1}^S = h(\{a^s\}_{s=1}^S )$
   \STATE \textbf{return} $\{i^s\}_{s=1}^S$
\end{algorithmic}
\end{algorithm}
 
 In short, the algorithm greedily multiplies the particle with index $t$ that contributes the most to maximizing $\mathcal{L}(x_{1:n})$. $C^+$ function measures how much is gained \emph{and} lost by incrementing the multiplicity of a particle. We introduced a mapping function, $h$, which partitions the multiplicities of the particles, $\{a^s\}_{s=1}^S$ to particle indices in a natural way.
 
 The KL reshuffling can be implemented using heap data structures so that it runs in $O(S \log S)$ time.

 \subsection{Likelihood Based KL Reshuffling}
 In addition to the importance weight based approach, we propose a reshuffling method based on the particle likelihoods, $\klp$. This will turn out to provide an excellent alternative to an ML based search heuristic.  
 
 Here, we focus on the KL divergence between the particle distribution and the posterior distribution ($r(x)=p(x_{1:n}|y_{1:n})$).
 Minimizing the expression in Equation \ref{eq:KL} involves handling the intractable distribution, so we approach it from another angle. Since the marginal log-likelihood is fixed and the KL divergence is non-negative, we may instead maximize the ELBO \cite{jordan1999introduction}, i.e., 

\begin{equation} 
\label{eq:elbo_1}
\begin{split}
    \mathcal{L}(x_{1:n}) &=\log p(y_{1:n})-\text{KL}(q(x_{1:n})||p(x_{1:n}|y_{1:n})) \\ 
    &= \mathbb{E}_{q}\left[\log\frac{p(x_{1:n},y_{1:n})}{q(x_{1:n})}\right].
\end{split}
\end{equation}
 Consequently, minimizing the KL divergence is equivalent to maximizing the ELBO. By plugging in the particle distribution from Equation \ref{eq:particle_dist} and later Equation \ref{eq:multpl_distr} into  the ELBO in Equation \ref{eq:elbo_1}, we obtain
 
\begin{equation}
\label{eq:elbo}
\begin{split}
 \mathcal{L}(x_{1:n}) 
    &= \sum_{s=1}^S q(x_{1:n}^s) \log\frac{p(x^s_{1:n},y _{1:n})}{q(x^s_{1:n})}\\
    &= \sum_{s=1}^S \ddot q(x_{1:n}^s) \log\frac{p(x^s_{1:n},y _{1:n})}{\ddot q(x^s_{1:n})}\\
    &= \sum_{s=1}^S \frac{a^s}{S} \log\frac{p(x^s_{1:n},y _{1:n})}{\frac{a^s}{S}}\\
    &\propto \sum_{s=1}^S a^s \log\frac{p(x^s_{1:n},y _{1:n})}{a^s}.
\end{split}
\end{equation}
 
This equation implies that we can minimize the KL divergence from $q(x_{1:n})$ to $p(x_{1:n}|y_{1:n})$ by finding the $\ddot q^*$ that maximizes $\mathcal{L}(x_{1:n})$.  Also $\klp$ relies on Algorithm \ref{alg:kl_reshuffling}; the only difference is the input of the algorithm,  $u^s = p(x_{1:n}^s, y_{1:n}), \forall s$. 

\section{Total Variation Reshuffling}
\label{sec:tv_reshuffling}

The TV distance between two particle distributions is defined as

\begin{equation}
    \begin{aligned}
        \text{TV}(q, p) &= \frac{1}{2}\sum_{x\in \mathcal{Q}\cup\mathcal{P}} |q(x) - p(x)|\\
    \end{aligned}
\end{equation}

where $\mathcal{Q}=\{x:q(x)>0\}$ and $\mathcal{P}=\{x:p(x)>0\}$. The range of TV distance is in $[0, 1]$, zero when the distributions are identical and one when the distributions have disjoint supports.

\subsection{Weight Based TV Reshuffling}
\label{sec:TVrs}
Recall the multiplicity set $\mathcal{A}^S$ from Equation \ref{eq:multiplicity_set}, and let $\left(\mathcal{X}^n\right)^S$ be the support of $q$. We want to find the $\ddot q^*$ that minimizes the TV distance

\begin{equation}
\begin{split}
    \text{TV}\left(q, \ddot q\right) &= \frac{1}{2}\sum_{x_{1:n}\in\left(\mathcal{X}^n\right)^S} |q(x_{1:n}) - \ddot q(x_{1:n}) |\\ &=\frac{1}{2}\sum_s |w^s - \ddot w^s |\\
    &=\frac{1}{2} \sum_s |w^s - \frac{a^s}{S}|,
\end{split}
\end{equation}
which is equivalent to finding $\{a^{*s}\}_{s=1}^S\in \mathcal{A}^s$ that minimizes 
$\frac{1}{2}\sum_s |w^s S - a^s|$. Let 

\begin{equation}
    \alpha^s = w^s S - \floor*{w^sS}, \quad\alpha =\sum_s \alpha^s
\end{equation}

and note that $\alpha\in\mathbb{Z}^+_0$, since $\sum_s w^sS \in\mathbb{Z}^+_0$ and $\sum_s \floor*{w^sS}\in\mathbb{Z}^+_0$.

Finally, assume that $\alpha^1\geq\ldots\geq\alpha^s$, and let $\ddot g$ be the greedy rounding of $w$. Computing $\ddot g$ according to  Algorithm \ref{alg:tv_reshuffling}, we claim that $\{\ddot g^s\}_{s=1}^S = \{a^{*s}\}_{s=1}^S$ and prove it in the Supplementary Materials. When the inputs in Algorithm \ref{alg:tv_reshuffling} are the importance weights, $u^s = w^s$ for all $s$, the resulting reshuffling algorithm is TV$_w$. Similar to the Algorithm \ref{alg:kl_reshuffling}, $h$ function maps the ancestor multiplicities to particle indices.

\begin{algorithm}[t]
   \caption{Total Variation Reshuffling}
   \label{alg:tv_reshuffling}
\begin{algorithmic}
   \STATE {\bfseries Input:} $\{u^s\}_{s=1}^S$
   \STATE {\bfseries Output:} $\{i^s\}_{s=1}^S$
   \STATE set $a^s = u^s S$, $\forall s$ 
   \STATE set $\alpha^s = a^s - \floor*{a^s}$, $\forall s$ 
   \STATE compute $\alpha = \sum_s \alpha^s$
   \STATE order $\alpha^1 \geq \ldots \geq \alpha^S$

   \FOR{$s=1,\ldots,S$}
   \IF{$s \leq \alpha$}
   \STATE set $\ddot g^s = \ceil*{a^s}$
   \ELSE
   \STATE set $\ddot g^s = \floor{a^s}$
   \ENDIF
   \ENDFOR
   \STATE map $\{i^s\}_{s=1}^S = h( \{\ddot g^s\}_{s=1}^S )$
   \STATE \textbf{return} $\{i^s\}_{s=1}^S$
\end{algorithmic}
\end{algorithm}

The TV reshuffling can be implemented to run in $O(S \log S)$ time using any sorting algorithm or in time $O(S)$ using radix sort  with a  negligible loss of numeric precision.

\begin{figure*}[t]
\centering
\subfigure[$L_{0-1}$ loss]{%
\label{fig:l01_sv}%
\includegraphics[width=0.25\textwidth]{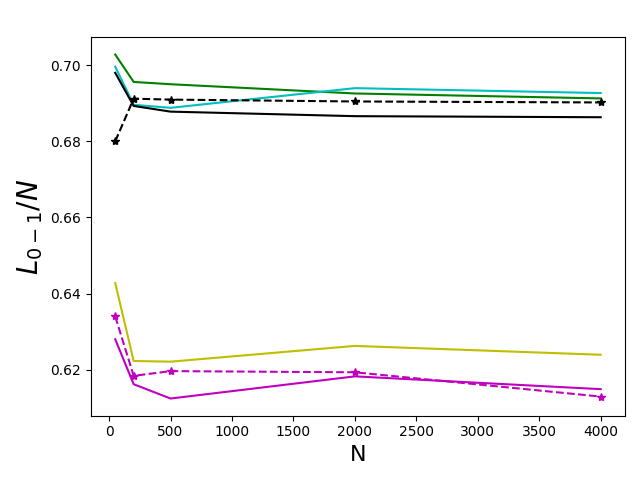}}%
\hskip 0.3cm
\subfigure[$L_1$ loss]{%
\label{fig:l1_sv}%
\includegraphics[width=0.25\textwidth]{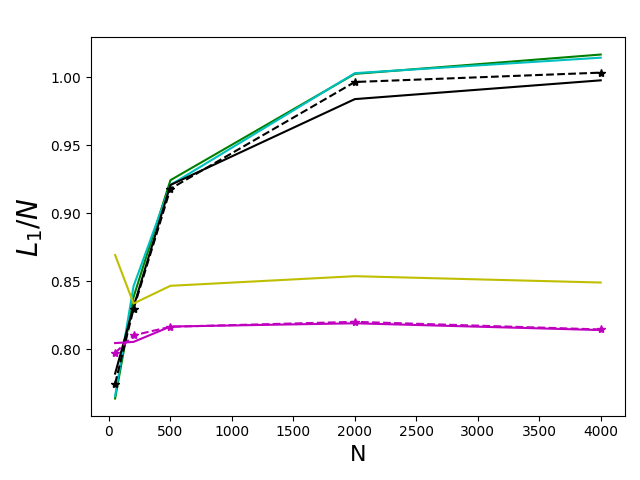}}%
\hskip 0.3cm
\subfigure[$L_2$ loss]{%
\label{fig:l2_sv}%
\includegraphics[width=0.25\textwidth]{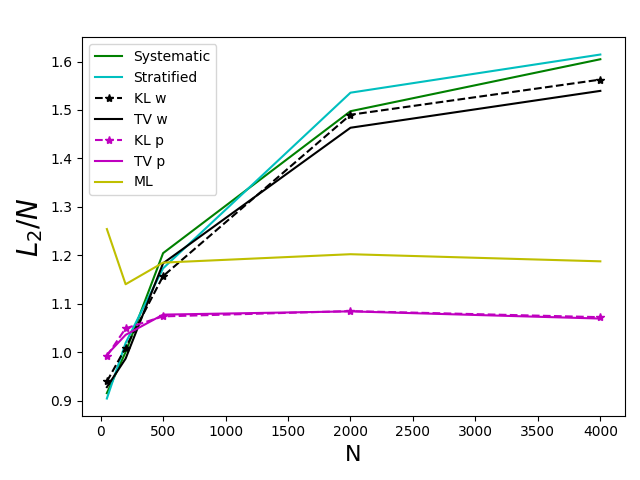}}%
\caption{\textbf{Bayesian estimators:} Average losses per $N$ on the SV model when $S=500$. x-axes are the varying number of $N$ whereas the y-axes are the average loss.}
\label{fig:SV_loss_S500}
\vskip -0.2in
\end{figure*}

\subsection{Likelihood Based TV Reshuffling}
TV reshuffling can be applied to $p(x_{1:n}, y_{1:n})$ and, also this, yields an alternative to an ML search heuristic. That is, we can use the TV reshuffling  to minimize the TV distance between $q(x_{1:n})$ and $p(x_{1:n}, y_{1:n})$ yielding TV$_p$ – TV reshuffling with likelihoods. Similar to the KL reshuffling, $\tvp$ requires the input of the reshuffling algorithm to be the particle likelihoods, rather than the importance weights. 

In Sections \ref{sec:kl_reshuffling} and \ref{sec:tv_reshuffling}, we introduced a class of reshuffling algorithms that perform offspring selection by minimizing separate statistical distances with respect to the importance weights, $\klw$ and $\tvw$. Both of the weight based methods are off-the-shelf reshuffling schemes and can be applied to the standard PF algorithm in Algorithm \ref{alg:bpf} without any workaround.

The likelihood based reshuffling schemes; $\klp$ and $\tvp$ work with the particle likelihoods, therefore we use them in the updated PF algorithm described in Section \ref{sec:bpf_likelihood}.

\section{Experiments} 
In order to compare the performance of our statistical distance based offspring selection schemes with the stratified and systematic resampling, we use a variety of loss functions (binary loss ($L_{0-1}$), absolute loss ($L_1$) and quadratic loss ($L_2$)) and their corresponding estimators (maximum a posteriori, minimum mean absolute error and minimum mean squared error estimators) \cite{murphy2012machine}. In addition to the mentioned Bayesian estimators, we also use \emph{sampled} estimators where we simply sample a particle trajectory with respect to its importance weight.  

We perform two sets of experiments; SMC (using BPF) and pMCMC (using particle Gibbs, PG \cite{pmcmc}). We use two different models and compare the offspring selection schemes in terms of losses, particle degeneracy, parameter estimation, and auto-correlations. 

\subsection{SMC Experiments}

Stochastic volatility (SV) and non-linear (NL) non-Gaussian state space models are used for the SMC experiments. In all the experiments, the number of particles is set to $S=500$ and the results are averaged over $50$ runs unless stated otherwise. 

\subsubsection{Stochastic Volatility Model}

The SV model is popular in real life cases of financial econometrics \cite{pmcmc}. Here $Y_{1:N}$ is a sequence of logarithmic returns\footnote{The logarithmic return at time $n$ is given by $y_{n} = \log (\frac{r_{n}}{r_{n-1}})$, where $r_n$ is the asset price at time $n$.} of asset prices $R_{1:N}$, and its volatility is believed to be governed by the latent process $X_{1:N}$. The model is given by

\begin{equation*}
    \mu(x_1) = \mathcal{N}(0,  \sigma^2 / (1 - \phi^2))
\end{equation*}
\begin{equation*}
    X_n = \phi X_{n-1} + \sigma V_n
\end{equation*}
\begin{equation*}
    Y_n = \beta\exp(X_n/2)E_n,
\end{equation*}
where $E_n$ and $V_n$ are random variables following the standard Gaussian distribution. In our experiments, we assumed the hyper-parameters are given as $(\sigma,\beta,\phi)=(1,0.5,0.91)$, the same setup as used in \cite{doucet2009tutorial}.

\begin{figure}[t]
\centering
\includegraphics[width=0.85\columnwidth]{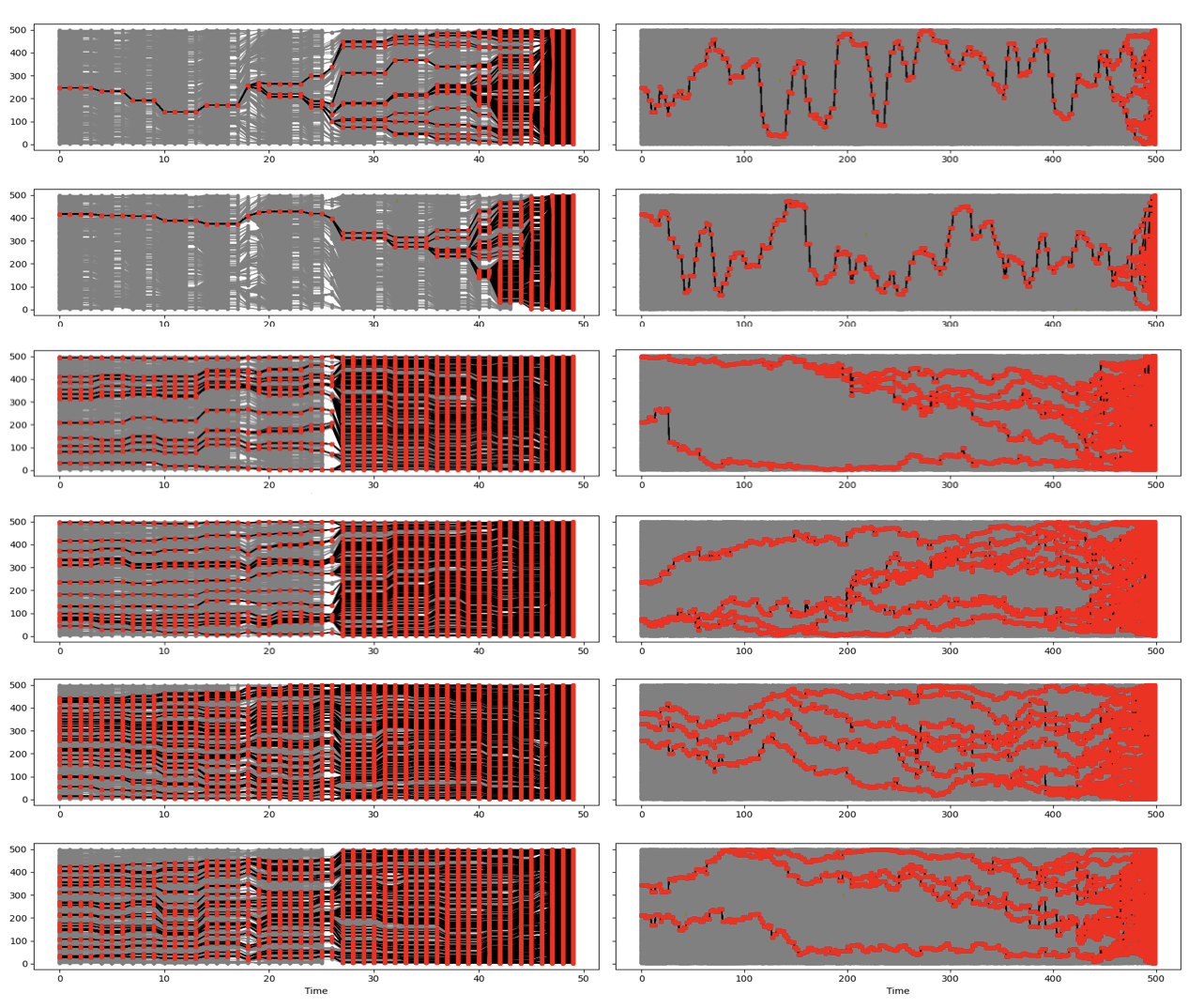}
\caption{From left to right, the plots depict particle degeneracy when $N\in\{50, 500\}$. The rows correspond to the offspring selection schemes (top to bottom): $\klp$, $\tvp$, $\klw$, $\tvw$,  systematic and stratified.
x-axes show the time $n\in N$ and y-axes indicate particle indices. The grey lines show how the particles are propagated. The black lines with red dots show the final particles' trajectories.}
\label{fig:particle_degeneracy}
\vskip -0.2in
\end{figure}

\begin{figure}[b]
\centering
\includegraphics[width=0.27\textwidth]{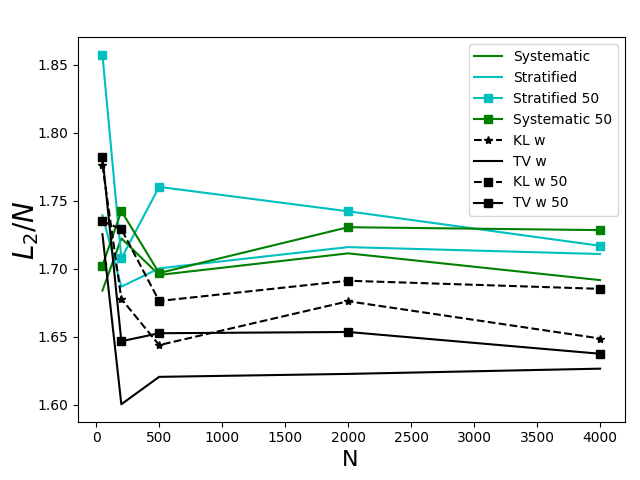}
\caption{\textbf{Sampled estimators:} Average $L_2$ loss per $N$ for weight based selection schemes in the SV model with different number of particles. The lines with square markers represent $S=50$ particles, and the rest represents $S=500$ particles. } 
\label{fig:l2_w_sampled_sv}%
\vskip -0.2in
\end{figure}

\begin{figure}[b]
\centering
\subfigure[Inferred states with ML]{%
\label{fig:nl_ml_infd}%
\includegraphics[width=0.23\textwidth]{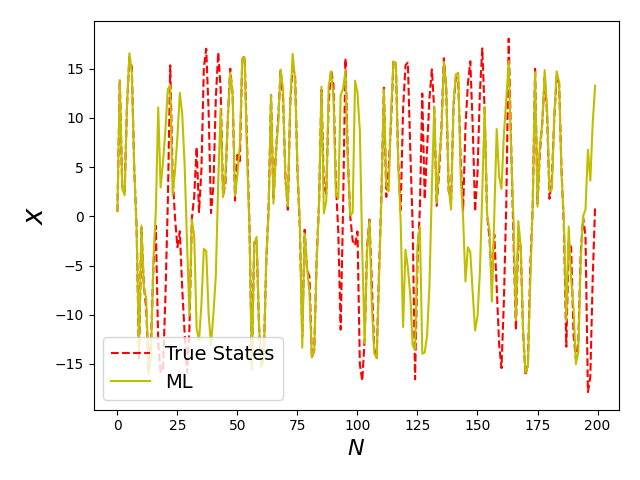}}%
\hskip 0.3cm
\subfigure[Inferred states with $\tvp$]{%
\label{fig:nl_tvw_infd}%
\includegraphics[width=0.23\textwidth]{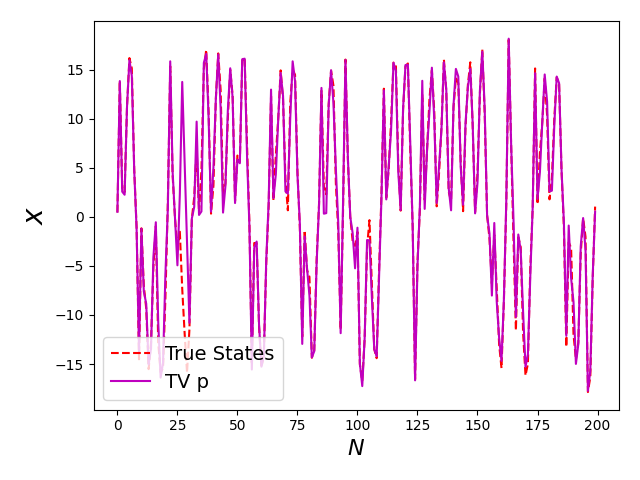}}%
\caption{Inferred states in the NL experiment. $\theta=(1, 1)$. The multimodality of the hidden state distribution makes (a) ML unequipped to solve it, in contrast to (b) $\tvp$. True states are represented by the red dashed line.}
\label{fig:nl_infd_ml_tv}
\vskip -0.2in
\end{figure}

Figure \ref{fig:SV_loss_S500} shows the average losses per time ($L/N$) of the Bayesian estimators using different offspring selection schemes. The likelihood based methods; ML, $\klp$ and $\tvp$, achieve lower loss than the weight based methods; especially when $N$ is large. Also, the weight based methods' $L_{0-1}$ losses are $N$-invariant, in contrast to its $L_1$ and $L_2$ losses. This behaviour can be explained by the Bayesian estimator of the $L_{0-1}$ loss, the maximum a posteriori estimator. The estimator is based on a single trajectory, belonging to the mode of $q(x_{1:N})$. 


\begin{figure*}[t]
\centering
\subfigure[$L_1$ loss, $\theta_1=(1,1)$]{%
\label{fig:nl_theta_11_L1}%
\includegraphics[width=0.23\textwidth]{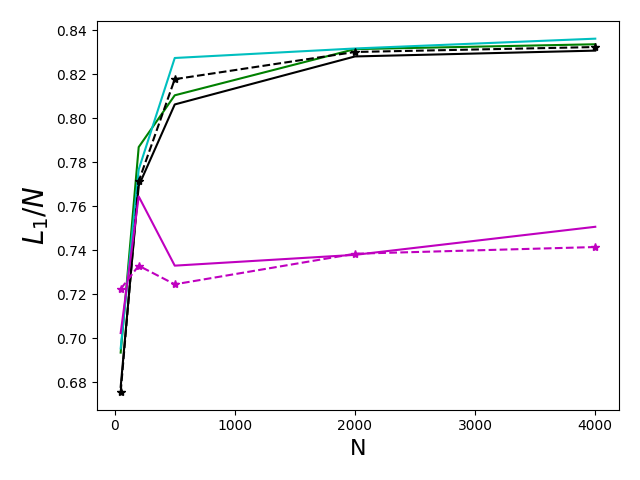}}%
\hskip 0.3cm
\subfigure[$L_1$ loss, $\theta_2=(10, 10)$]{%
\label{fig:nl_theta_1010_l1}%
\includegraphics[width=0.23\textwidth]{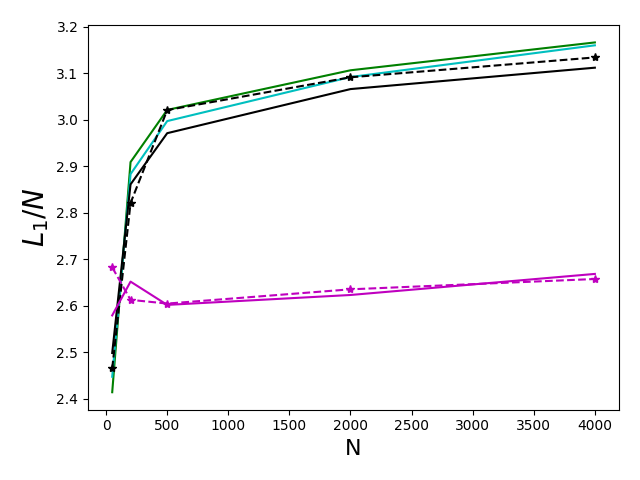}}%
\hskip 0.3cm
\subfigure[$L_2$ loss, $\theta_1=(1,1)$]{%
\label{fig:nl_theta_11_L2}%
\includegraphics[width=0.23\textwidth]{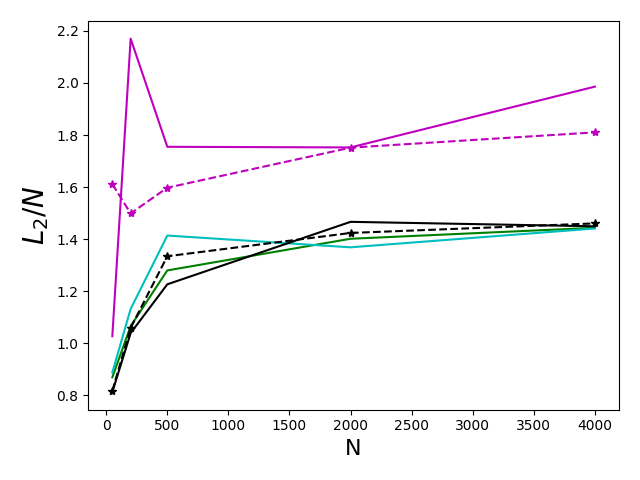}}%
\hskip 0.3cm
\subfigure[$L_2$ loss, $\theta_2=(10, 10)$]{%
\label{fig:nl_theta_1010_l2}%
\includegraphics[width=0.23\textwidth]{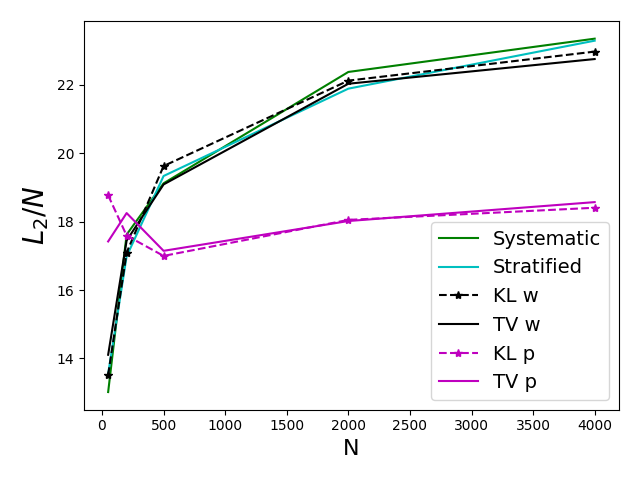}}%
\caption{\textbf{Bayesian estimators:}
Average losses per time for different $\theta$ values in the NL experiment.}
\label{fig:nl_l1_grid}
\vskip -0.2in
\end{figure*}

We show the particle degeneracy of each offspring selection scheme in Figure \ref{fig:particle_degeneracy}. The plots are generated using a single, $N=500$ long observation sequence from the set of SV sequences. 
The ML is not included, since it degenerates as soon as the resampling is done. From the $N=50$ plots (left column), it can be seen that the likelihood based methods ($\klp$ and $\tvp$) degenerate faster, and the final particles share a single common ancestor. The proposed weight based methods ($\klw$ and $\tvw$) degenerate less than the likelihood based methods and more than systematic and stratified resampling. The pattern is clearer when $N=500$. Importantly, all interesting PFs degenerate eventually, regardless of the resampling scheme. As such, and given the results of our likelihood based methods, one should reflect on whether a PF that degenerates necessarily is less useful than one that does not. 

In Figure \ref{fig:l2_w_sampled_sv}, the average $L_2$ loss per time of the sampled estimators of weight based selection schemes are shown. Here, we focus on weight based approaches, but more results can be found in the Supplementary Materials. Remarkably, as the figure shows, for all $N>50$ and with $S=500$, the PF based estimators that build on  $\tvw$ and $\klw$  only need 50 particles in order to obtain a lower $L_2$ loss than  that obtained by those building on stratified or systematic resampling.

\subsubsection{Non-linear Non-Gaussian State Space Model}

The NL non-Gaussian state space model is an important and often turned to example in the SMC literature \cite{kitagawa96, godsill2004monte, pmcmc,smith2013sequential}. It is an interesting test for PFs, as $p(x_{1:n}|y_{1:n})$ is multimodal \cite{godsill2004monte}. In order to excel in this experiment, the PF has to have particles in the vicinity of these modes, i.e. distribute its particles into multiple clusters in the latent space. The model is defined as

\begin{equation*}
    X_1 \sim \mathcal{N}(0, \sigma^2_x)
\end{equation*}
\begin{equation*}
    X_n = \frac{X_{n-1}}{2} + 25 \frac{X_{n-1}}{1 + X_{n-1}^2} + 8\cos(1.2n) + V_n
\end{equation*}
\begin{equation*}
    Y_n = \frac{X_n^2}{20} + U_n
\end{equation*}

where $V_n\sim\mathcal{N}(0, \sigma^2_x)$ and $U_n \sim \mathcal{N}(0,\sigma^2_y)$.

\setcounter{figure}{5}
\begin{figure*}[hb]
\centering
\subfigure[KL$_w$ $N=20, S=100$]{%
\label{fig:acf_kl_w_T20}%
\includegraphics[width=0.23\textwidth]{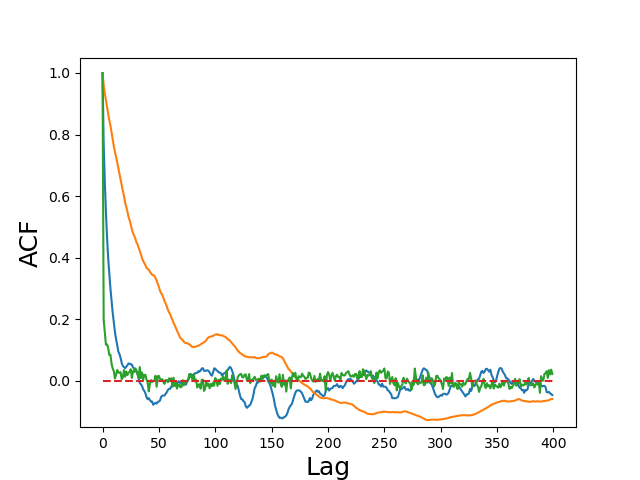}}%
\hskip 0.3cm
\subfigure[Strat. $N=20, S=100$]{%
\label{fig:acf_stratified_T20}%
\includegraphics[width=0.23\textwidth]{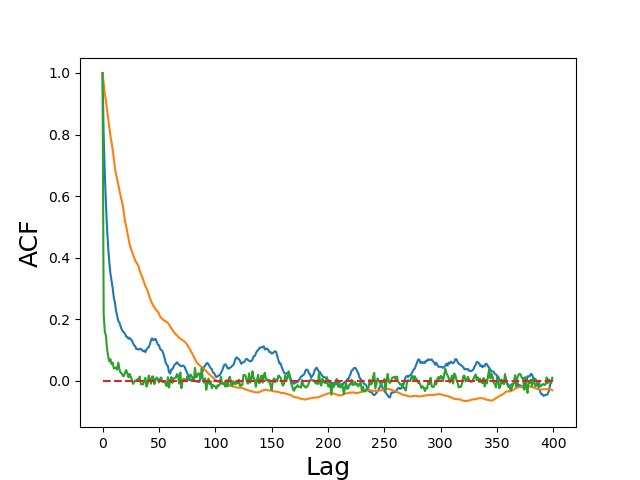}}%
\subfigure[KL$_w$ $N=50, S=100$]{%
\label{fig:acf_kl_w_T50}%
\includegraphics[width=0.23\textwidth]{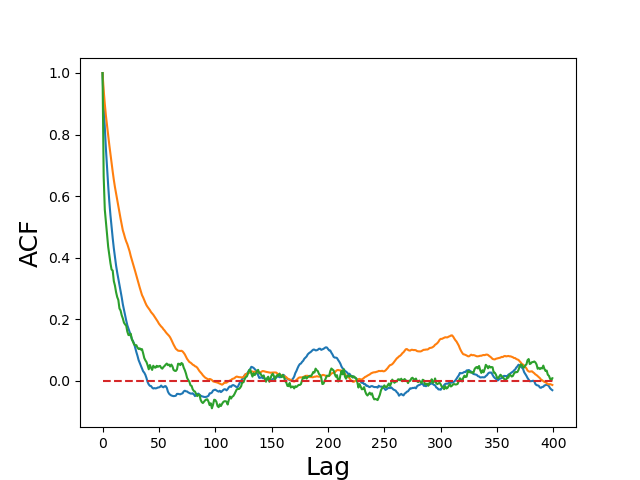}}%
\hskip 0.3cm
\subfigure[Strat. $N=50, S=100$]{%
\label{fig:acf_stratified_T50}%
\includegraphics[width=0.23\textwidth]{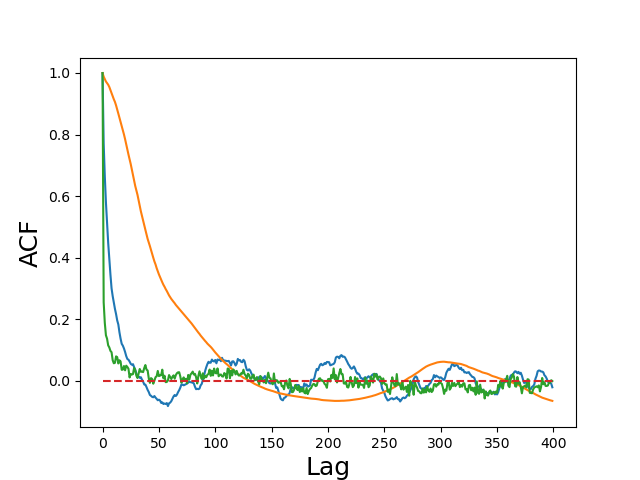}}%
\\
\subfigure[KL$_w$ $N=100, S=100$]{%
\label{fig:acf_kl_w_T100}%
\includegraphics[width=0.23\textwidth]{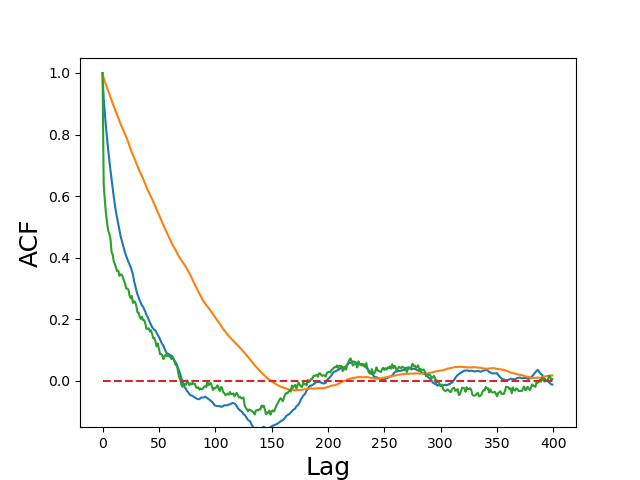}}%
\hskip 0.3cm
\subfigure[Strat. $N=100, S=100$]{%
\label{fig:acf_stratified_T100}%
\includegraphics[width=0.23\textwidth]{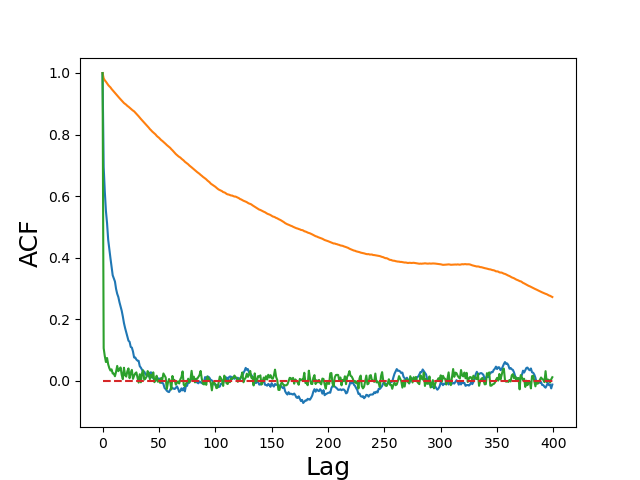}}%
\subfigure[KL$_w$ $N=500, S=100$]{%
\label{fig:acf_kl_w_T500}%
\includegraphics[width=0.23\textwidth]{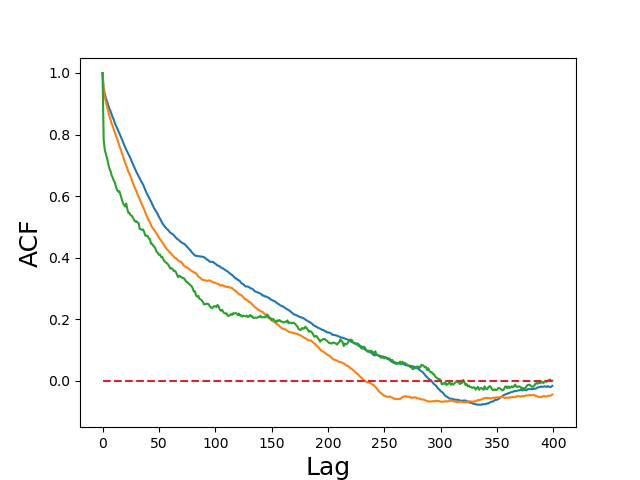}}%
\hskip 0.3cm
\subfigure[Strat. $N=500, S=100$]{%
\label{fig:acf_stratified_T500}%
\includegraphics[width=0.23\textwidth]{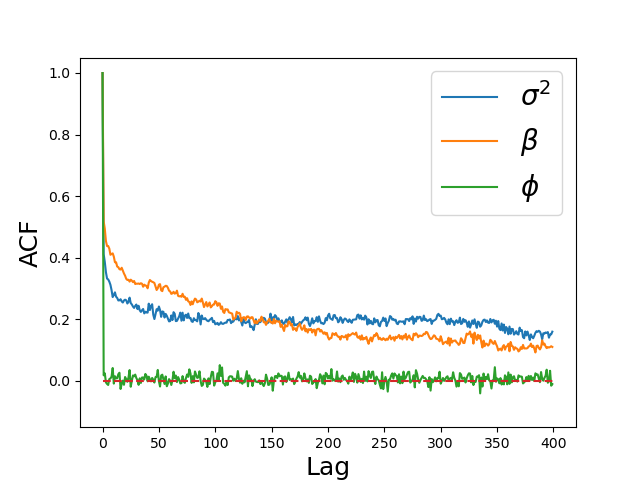}}%
\\
\caption{ACF plots of $\klw$ and stratified resampling for PG experiments.}
\label{fig:kl_w_stratified_acf}
\vskip -0.2in
\end{figure*}

In this group of experiments, we investigate the effects of the parameters $\theta=(\sigma^2_x, \sigma_y^2)$. Observations of varying lengths $N$ are generated with $\theta_1=(1, 1)$ and $\theta_2=(10, 10)$. More experiments with different parameters are available in the Supplementary Materials. 

As revealed by the experiment illustrated in Figure \ref{fig:nl_infd_ml_tv}, the ML selection scheme infers states poorly in cases where the true posterior is multimodal. This is natural since  ML selects the particle with the highest likelihood, $p(x_{1:n}^s, y_{1:n})$, making the PF distribute all its particles centered at a single mode. Given this severe shortcoming of the ML reshuffling scheme, it will not be considered in the coming experiments. In contrast, both of our likelihood based methods $\tvp$ (in Figure \ref{fig:nl_tvw_infd}) are $\klp$ (not shown here) able to account for the multimodality.

The performance of Bayesian estimators for different parameter settings are shown in Figure \ref{fig:nl_l1_grid}. We make the following two observations based on the plots. First, it is important to consider multiple loss functions. $L_1$ loss plots (Figures \ref{fig:nl_theta_11_L1} and \ref{fig:nl_theta_1010_l1}) indicate the likelihood based schemes provide the best estimates, whereas Figure \ref{fig:nl_theta_11_L2} show this is not always the case. Second, our weight based methods obtain smaller losses in three out of the four cases.

\subsection{pMCMC Experiments}
Here we will present results when, again, considering the SV model, now parameterized by the unknown model parameters $\theta=(\sigma, \beta, \phi)$. Before applying the method to real data, we first consider data generated from the model used in the SMC experiments above, i.e., the true model parameters were $\theta=(\sigma, \beta, \phi)=(1, 0.5, 0.91)$. This allows us to more precisely evaluate the performances of the methods. 

The main objective of the following experiments is to evaluate the mixing and parameter estimation capabilities of the PG when using deterministic offspring selection. We use the auto-correlation functions (ACF) to quantify the correlation between samples in the chain as a function of lag. In order to simplify our analysis, we constrain ourselves to comparing with only the stratified resampling scheme.

Inspired by the priors chosen in \cite{ancestor_sampling}, we assumed that all parameters were independent, while assigning $\sigma$ and $\beta$ inverse Gamma priors (both the shape and rate parameters are set to $0.001$). Meanwhile, the posterior over $\phi$ does not admit a closed-form expression, we modelled $\phi$ as in \cite{ancestor_sampling} and it was approximated using a rejection sampler \cite{kim1998stochastic, ancestor_sampling}.  See Supplementary Materials for details and posterior derivations.

Table \ref{tab:pg_est} shows the parameters estimated by $\klw$ and stratified resampling with varying time. The estimates are calculated by taking the median over the last $5,000$ samples. The closest estimates to the true model parameters are shown in bold. 
For larger $N$, $\klw$ achieves closer parameter estimates. For $N=500$, we observe that the stratified resampling's $\beta$ estimate diverges dramatically. 

Figure \ref{fig:kl_w_stratified_acf} shows the corresponding ACFs. $\klw$ has strong mixing for small $N$.
Stratified resampling achieves lower auto-correlation values for smaller lags when $N\in \{20, 50\}$; however, it needs larger lag to obtain good mixing when $N\in \{100, 500\}$. One possible explanation why stratified immediately gets uncorrelated samples for $\phi$ when $N=500$ is due to $g(y_n|x_n)$ in the SV model assigning low probability to all $x_{1:N}$ when $\beta$ is large (as is the case for stratified, see Table \ref{tab:pg_est}). As such the PG is sampling trajectories from a flat distribution, making them varied between iterations. The $\phi$ parameter's rejection sampler mainly depends on the sampled trajectory (both in terms of expected value and variance), and so, because of the dissimilarity between them, the sampler draws samples seemingly independent of its previous sample.

\begin{table}[t]
\caption{PG parameter estimation results} 
\label{tab:pg_est}
\vskip 0.1in
\begin{center}
\begin{small}
\begin{sc}
\begin{tabular}{c|ccc|ccc}
\toprule
\multicolumn{1}{c}{} & \multicolumn{3}{c}{$\klw$} & \multicolumn{3}{c}{Stratified} \\
\midrule
$N$      & $\sigma^2$ & $\beta$ & $\phi$    & $\sigma^2$ & $\beta$ & $\phi$ \\
\midrule
$20$     & $1.42$ & $0.36$ & $0.83$      & $\mathbf{1.37}$ & $\mathbf{0.39}$  & $\mathbf{0.84}$ \\
$50$     & $\mathbf{0.72}$ & $\mathbf{0.49}$ & $0.88$      & $1.77$ & $0.14$  & $\mathbf{0.89}$ \\
$100$    & $\mathbf{0.69}$ & $\mathbf{0.55}$ & $\mathbf{0.94}$      & $2.05$ & $0.42$  & $0.86$ \\
$500$    & $\mathbf{0.56}$ & $\mathbf{0.38}$ & $\mathbf{0.92}$      & $2.11$ & $13.09$ & $0.95$ \\ 
\midrule
True     & $1.00$ & $0.50$ & $0.91$               & $1.00$ & $0.50$ & $0.91$ \\
\bottomrule
\end{tabular}
\end{sc}
\end{small}
\end{center}
\vskip -0.3in
\end{table}

Unfortunately, the likelihood based reshuffling schemes converge almost immediately to small variance values, far from the true variances. In short, this occurs since these methods reshuffle, partly, based on the transition probability, which in the SV model obtains extremely high likelihood for the trajectories close to its mean when $\sigma^2<1$. In the Supplementary Materials, we further discuss this, and provide plots; however, in the rest of the pMCMC experiments we shall not consider the likelihood based methods.

Finally, we test the performance of our proposed weight based offspring selection methods using Standard and Poor's (S\&P) 500 dataset \cite{real_data}. For the experiments, we extract the data between 2006-04-03 and 2014-03-28, corresponding to $N=2,011$ as in \cite{ancestor_sampling}.

\begin{figure}[t]
\centering
\subfigure[$\tvw$, $S=20$]{%
\label{fig:sp_2011_2}%
\includegraphics[width=0.23\textwidth]{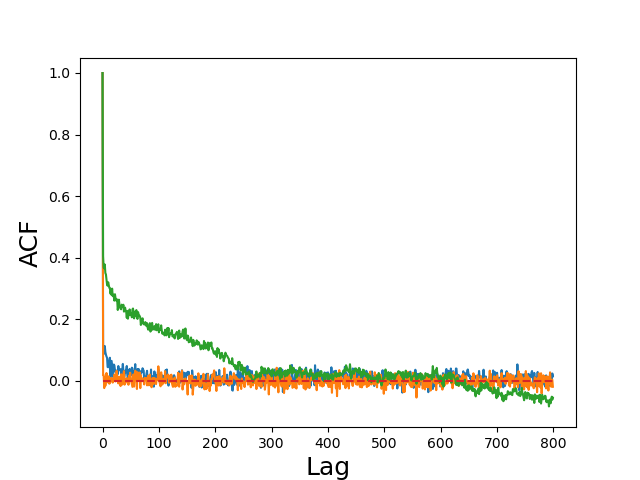}}%
\subfigure[Stratified, $S=20$]{%
\label{fig:sp_2011_3}%
\includegraphics[width=0.23\textwidth]{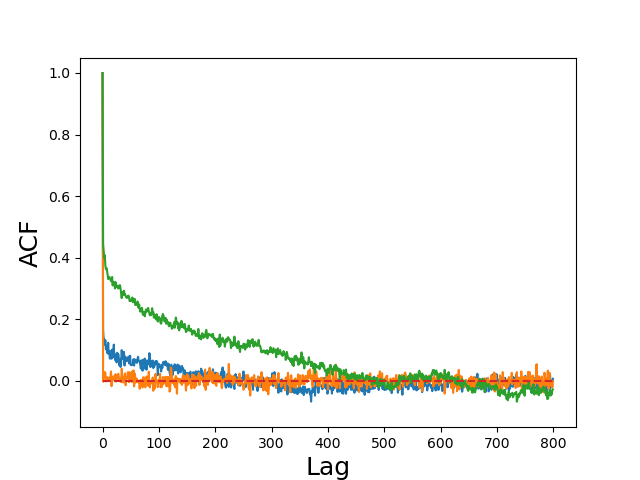}}%
\\
\subfigure[$\tvw$, $S=50$]{%
\label{fig:sp_2011_5}%
\includegraphics[width=0.23\textwidth]{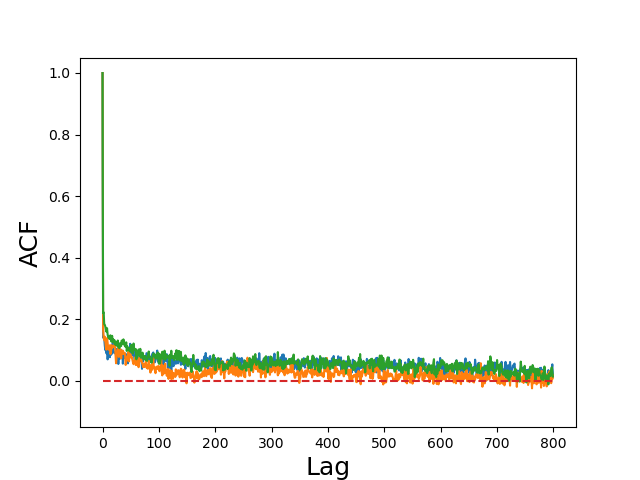}}%
\subfigure[Stratified, $S=50$]{%
\label{fig:sp_2011_6}%
\includegraphics[width=0.23\textwidth]{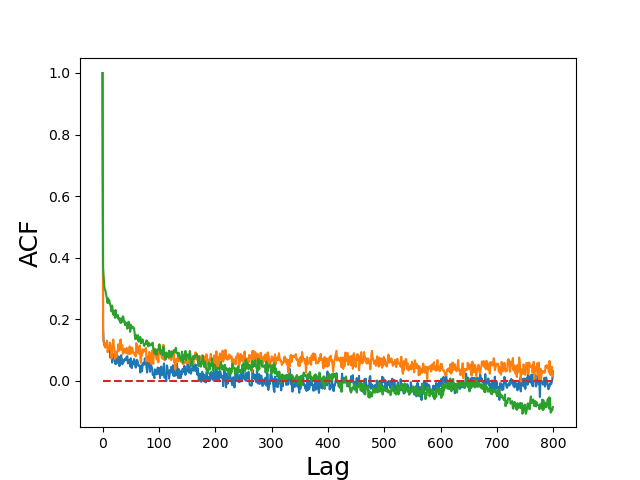}}%
\\
\subfigure[$\tvw$, $S=100$]{%
\label{fig:sp_2011_8}%
\includegraphics[width=0.23\textwidth]{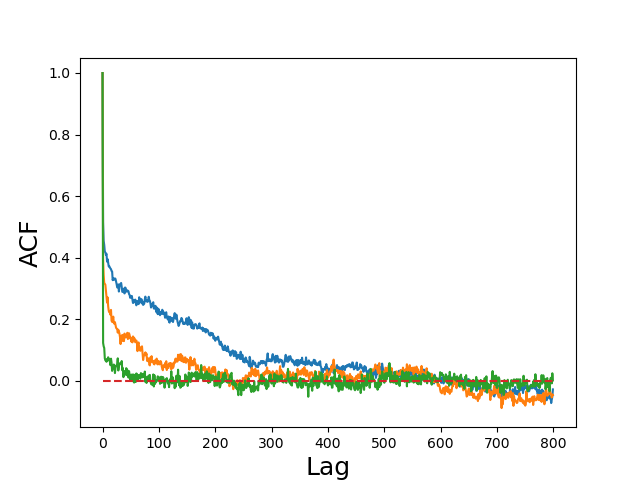}}%
\subfigure[Stratified, $S=100$]{%
\label{fig:sp_2011_9}%
\includegraphics[width=0.23\textwidth]{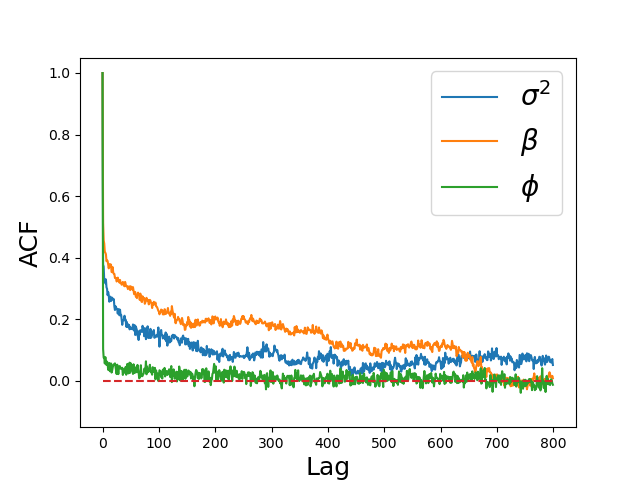}}%

\caption{ACF plots of $\tvw$ and stratified resampling for S\&P $N=2,011$ experiments. The columns are $\tvw$ (left) and stratified (right). The rows are the number of particles $S \in \{20, 50, 100\}$, respectively.}
\label{fig:real_2011_acf}
\vskip -0.2in
\end{figure}

We investigate how the resampling methods work for varying number of particles. Figure \ref{fig:real_2011_acf} shows the ACFs of $\tvw$ and stratified resampling for $N=2,011$ data, based on the last $5,000$ iterations (see the Supplementary Materials for $\klw$ results). When PG is trained with $20$ particles, both $\tvw$ and stratified resampling have low mixing for $\phi$. They perform on par for $\sigma^2$ and $\beta$. Both methods achieve good mixing for $\phi$ when more particles are used ($S=100$); however, stratified resampling suffers more for $\sigma^2$ and $\beta$, even when the lag is large. 

\section{Conclusion}
We introduce two novel offspring selection schemes that are based on minimizing a statistical distance between the normalized particle distribution prior to the offspring selection and the unweighted particle distribution obtained. The
first algorithm minimizes the Kullback-Leibler divergence and the second algorithm total variation distance. Moreover, both are proved to provide optimal solutions. 

We investigated the performance of several offspring selection schemes by embedding them in SMC and pMCMC algorithms for the stochastic volatility model and the non-linear non-Gaussian state space model.
Both our novel  offspring selection schemes performed better, with respect to the appropriate loss functions, than the state-of-the-art probabilistic resampling schemes, stratified and systematic. Our methods obtain better estimates of parameters and attain less auto-correlation when used in pMCMC.
Surprisingly, our offspring selection schemes performed better than stratified and systematic even when we allowed stratified and systematic to use a multiplicative factor of 10 more particles. 

As a byproduct, we also obtained two heuristic schemes that instead minimize the statistical distance to the joint distribution. These two had in several cases a desirable performance and we consider them, in particular, to be excellent alternatives for problems that seem fit to a ML based heuristic search.

\section{Conclusion}
We introduce two novel offspring selection schemes that are based on minimizing a statistical distance between the normalized particle distribution prior to the offspring selection and the unweighted particle distribution obtained. The
first algorithm minimizes the Kullback-Leibler divergence and the second algorithm total variation distance. Moreover, both are proved to provide optimal solutions. 

We investigated the performance of several offspring selection schemes by embedding them in SMC and pMCMC algorithms for the stochastic volatility model and the non-linear non-Gaussian state space model.
Both our novel  offspring selection schemes performed better, with respect to the appropriate loss functions, than the state-of-the-art probabilistic resampling schemes, stratified and systematic. Our methods obtain better estimates of parameters and attain less auto-correlation when used in pMCMC.
Surprisingly, our offspring selection schemes performed better than stratified and systematic even when we allowed stratified and systematic to use a multiplicative factor of 10 more particles. 

As a byproduct, we also obtained two heuristic schemes that instead minimize the statistical distance to the joint distribution. These two had in several cases a desirable performance and we consider them, in particular, to be excellent alternatives for problems that seem fit to a ML based heuristic search. 

\section*{Acknowledgements}
This project is funded by the Swedish Foundation with Strategic Research grant BD15-0043 and the Swedish Research Council grant 2018-05417\_VR.  The computations and data handling were enabled by resources provided by the Swedish National Infrastructure for Computing (SNIC), partially funded by the Swedish Research Council through grant agreement no. 2018-05973.

\bibliography{main.bib}
\bibliographystyle{icml2021}

\newpage
\onecolumn

\begin{center}
\vspace{0.25in} 
\fontsize{14}{1}\textbf{Statistical Distance Based Deterministic Offspring Selection in SMC Methods \\ Appendix} \\
\vspace{0.1in} \rule{\textwidth}{1pt}
\end{center}
\setcounter{page}{1}

\appendix

\section{KL Reshuffling Proof}
\label{sup_sec:kl_proof}
In this section, we show that the KL reshuffling gives the optimal solution, $\ddot q^*(x_{1:n}) = \sum_s \frac{a^{*s}}{S}\delta(x_{1:n}^s)$, maximizing the objective function $\mathcal{L}$ (in the weight based approach, $ \mathcal{L}$ is the negative KL divergence from $\ddot q(x_{1:n})$ to $q(x_{1:n})$, whereas, in the likelihood based approach, $\mathcal{L}$ is the ELBO).

First, the inputs $u^s$ (in the weight based approach, $u^s = w^s$ and in the likelihood based approach, $u^s = p(x_{1:n}^s, y_{1:n})$) are ordered in descending order ($u^1\geq\dots\geq u^S$). A particle $s$ has multiplicity $a^s$. We define 
\begin{equation}
\begin{aligned}
    f(a^s, u^s) &= a^s \log \frac{u^s}{a^s} \\
    C^+(a^s, u^s) &= f(a^s+1, u^s) - f(a^s, u^s) \\
    C^-(a^s,u^s) &= f(a^s, u^s) - f(a^s-1, u^s) .
\end{aligned}
\label{sup_eq:c_def}
\end{equation}
where $f(a^s, u^s)$ is the contribution of particle $s$ to $\mathcal{L}$. $C^+$ and $C^-$ functions measure how much the $\mathcal{L}$ changes by increasing and decreasing $a^s$ by $1$. 

$\ddot q^*$ is the particle distribution maximizing $\mathcal{L}$, and let $\ddot q$ be the distribution obtained by doing KL reshuffling. We need to prove that $\ddot q = \ddot q^*$. 

Let $g=\argmax_s C^+(a^s, u^s)$ and $a^g$ the current multiplicity of particle $g$. Consider the first time KL reshuffling obtains an $a^g$ such that $a^g > a^{*g}$.
Also, let $o$ be the index of the leftmost particle such that $a^{*s}>a^{s}$. We will refer this particle as the \emph{optimal} particle, whose optimal multiplicity is $a^{*o}$. This follows the following statements:
    \begin{equation*}
    \begin{aligned}
        \textbf{(i) }& a^g = a^{*g} +1 \\
        \textbf{(ii) }& - C^{+}(a^{*g}, u^g) + C^{-}(a^{*o}, u^o) \geq 0 \\
        \textbf{(iii) }& C^{+}(a^{g}-1, u^g) - C^{+}(a^o, u^o) \geq 0 \\
        \textbf{(iv) }
        & C^{+}(a^{*g}, u^g) - C^{+}(a^o, u^o) \geq 0 
    \end{aligned}
    \end{equation*}
    
\begin{itemize}
    \item Item \textbf{(i)} reflects the current situation in the algorithm (recall ``Consider the first time KL reshuffling obtains an $a^g$ such that $a^g > a^{*g}$''). 
    \item Item \textbf{(ii)} states that the cost of changing the optimal solution is greater than or equal to zero.
    \item Item \textbf{(iii)} shows that it is more or equally beneficial to increase the multiplicity of particle $g$ from $a^g-1$ to $a^g$ than to increase the multiplicity of particle $o$ from $a^o-1$ to $a^o$. 
    \item Item \textbf{(iv)} states the same information as item \textbf{(iii)} since $a^g = a^{*g} +1$.
\end{itemize}
    
By combing the items \textbf{(ii)} and \textbf{(iv)}, we get
    \begin{equation}
    \label{eq:truth}
        C^{-}(a^{*o}, u^o) - C^{+}(a^o, u^o) \geq 0.
    \end{equation}
    
When we plug in the definitions of $C^{+}$ and $C^{-}$ in Equation \ref{sup_eq:c_def}, we get
    \begin{equation}
    \begin{aligned}
        C^{-}(a^{*o}, u^o) - C^{+}(a^o, u^o) &= f(a^{*o}, u^o) - f(a^{*o}-1, u^o) - f(a^o+1, u^o) + f(a^o, u^o)\\
        &= (a^{*o} - (a^{*o}-1)) \log u^o 
        - (a^o+1 - a^o) \log u^o\\
        &\qquad - a^{*o} \log a^{*o} + (a^{*o}-1) \log (a^{*o}-1) - a^o \log a^o + (a^o+1) \log (a^o+1) \\
        &= (a^{*o}-1) \log (a^{*o}-1) + (a^o+1) \log (a^o+1) - (a^{*o} \log a^{*o} + a^o \log a^o)
    \end{aligned}
    \end{equation}
    
    Let $Q = a^{*o} + a^o$. Then, the above equation can be written as
    \begin{equation}
    \label{eq:contradiction}
    \begin{aligned}
        C^{-}(a^{*o}, u^o) - C^{+}(a^o, u^o) &= Q \left[ \frac{(a^{*o}-1)}{Q} \log \frac{(a^{*o}-1) }{Q}+ \frac{(a^o+1)}{Q} \log \frac{(a^o+1) } {Q}
         - (\frac{a^{*o}}{Q} \log \frac{a^{*o}}{Q} + \frac{a^o}{Q} \log \frac{a^o}{Q}) \right]\\
        &= Q \left[ H\left( \frac{a^{*o}}{Q}, \frac{a^o}{Q} \right) - 
        H\left( \frac{a^{*o}-1}{Q}, \frac{a^o+1}{Q} \right) \right],
    \end{aligned}
    \end{equation}
    where $H(\cdot,\cdot)$ is the entropy of a categorical distribution with two categories.

    Recall that $a^{*o} >a^o$, and note that there are two cases. First, if $a^{*o} >a^o + 1$, then $ H\left( \frac{a^{*o}-1}{Q}, \frac{a^o+1}{Q} \right) >  H\left( \frac{a^{*o}}{Q}, \frac{a^o}{Q} \right)$, since the categorical distribution on the left-hand side is flatter than the one on the right-hand side, stating that $C^{-}(a^{*o}, u^o) - C^{+}(a^o, u^o) < 0$. This results in a contradiction between Equations \eqref{eq:truth} and \eqref{eq:contradiction}, and so this case cannot be true.
    
    The remaining case, $a^{*o} = a^o + 1$, leaves the entropy unchanged ($H\left( \frac{a^{*o}-1}{Q}, \frac{a^o+1}{Q} \right) =  H\left( \frac{a^{*o}}{Q}, \frac{a^o}{Q} \right)$), while both Equations \eqref{eq:truth} and \eqref{eq:contradiction} give $C^{-}(a^{*o}, u^o) = C^{+}(a^o, u^o)$.
    Using this equality in any of the items (\textbf{ii}-\textbf{iv}) tells us that incrementing $a^g$ was as beneficial as incrementing $a^{o}$ according to $C^+$, in which case we  increase the leftmost particle, either $o$ or $g$. 
    
    Importantly, there may be multiple optimal solutions, but we always choose the solution which is most skewed to the left.

\newpage
\section{TV Reshuffling Proof}
\label{sup_sec:tv_proof}

Here we prove that $\{\ddot g^s\}_{s=1}^S$ is indeed the optimal set of multiplicities, $\{\ddot a^{*s}\}_{s=1}^S$ minimizing the distance \footnote{This proof is for the weight based TV reshuffling. However, the same proof holds for the likelihood based version as well.}
\begin{equation}
\begin{aligned}
    \text{TV}(q,\ddot q) &= \frac{1}{2}\sum_{x_{1:n}\in\left(\mathcal{X}^n\right)^S} |q(x_{1:n}) - \ddot q(x_{1:n}) | \\
    &= \frac{1}{2} \sum_s | w^s - \ddot w^s | \\
    &= \frac{1}{2} \sum_s | w^s - \frac{\ddot g^s}{S} | \\
\end{aligned}
\label{eq:app_tv}
\end{equation}

Recall from the Section 4.1 that 
\begin{equation}
\begin{aligned}
    \alpha^s &= w^s S - \floor{w^s S} \\
             &= a^s - \floor{a^s} \\
    \alpha &= \sum^S_{s=1} \alpha^s
\end{aligned}
\end{equation}

We should note that $a^s$ values do not have to be integers since $a^s = w^s S$.

First, we sort the particles in descending order w.r.t. $\{\alpha^s\}_{s=1}^S$. Once the particles are sorted, $\alpha$ constitutes a decision boundary, deciding which particles should be rewarded with an additional offspring ($\ddot g^s = \ceil{a^s}$ if $s\leq\alpha$) and which should not ($\ddot g^s =\floor{a^s}$ if $s>\alpha$). 

Consider the case where $\ddot g \neq \ddot a^*$, i.e. our solution is not an optimal solution. There may be multiple optimal solutions, so assume that $\ddot a^* $ is the optimal solution which is the most similar (has the lowest TV distance) solution to $\ddot g$. Further, assume two conditions:

\begin{equation}
    \begin{aligned}
        \textbf{(i) }& \textnormal{$i$ is the first index such that $\ddot a^{*i} > \ddot g^i$} \\
        \textbf{(ii) }& \textnormal{$j$ is the last index such that $\ddot a^{*j}< \ddot g^j$} \\
    \end{aligned}
    \label{sup_eq:tv_conditions}
    \end{equation}
    
\begin{itemize}
    \item Item \textbf{(i)} indicates the optimal solution assigned more multiplicity to particle $i$.
    \item Item \textbf{(ii)} indicates the optimal solution assigned less multiplicity to particle $j$.
\end{itemize}

Since $\ddot a^*$ and $\ddot g$ are multiplicities, they are integers, making $|\ddot a^{*i} - \ddot g^i| \geq 1$ and $|\ddot a^{*j} - \ddot g^j| \geq 1$. 

There are four cases where particles $i$ and $j$ can be located based on $\{\alpha^s\}_{s=1}^S$ sorting and the arbitrary decision boundary ($\alpha$). Below, we will cover all these cases and show that $\ddot g$ is an optimal solution by contradiction. 

\begin{enumerate}
    \item \textbf{$i \leq \alpha$ and $j \leq \alpha$:} \\Both of the particles are placed at the left-hand side of the decision boundary. The TV reshuffling assigns the rounded up values as the multiplicities; $\ddot g^i = \ceil{a^i}$ and $\ddot g^j = \ceil{a^j}$. Combining these with the conditions in Equation \ref{sup_eq:tv_conditions}, we get $\ddot a^{*i} > \ddot g^i \geq a^i$ and $\ddot a^{*j} < a^j \leq \ddot g^j$. By changing $\ddot a^*$ (transferring $1$ multiplicity from $i$ to $j$), one obtains a closer solution to $\ddot g$, which is a better than or as good to the ``optimal'' solution. Either case, there is a contradiction. 
    \item \textbf{$i \leq \alpha$ and $j > \alpha$:} \\$i$ is placed at the left-hand side and $j$ is placed at the right-hand side of the decision boundary. The TV reshuffling assigns the multiplicities $\ddot g^i = \ceil{a^i}$ and $\ddot g^j = \floor{a^j}$. Combining these with the conditions in Equation \ref{sup_eq:tv_conditions}, we get $\ddot a^{*i} > \ddot g^i \geq a^i$ and $\ddot a^{*j} < \ddot g^j \leq a^j$. By changing $\ddot a^*$ (transferring $1$ multiplicity from $i$ to $j$), one obtains a closer solution to $\ddot g$, which is a better than or as good as the ``optimal'' solution. In either case, there is a contradiction. 
    \item \textbf{$i > \alpha$ and $j \leq \alpha$:} \\$i$ is placed at the right-hand side and $j$ is placed at the left-hand side of the decision boundary. The TV reshuffling assigns the multiplicities $\ddot g^i = \floor{a^i}$ and $\ddot g^j = \ceil{a^j}$. Combining these with the conditions in Equation \ref{sup_eq:tv_conditions}, we get $\ddot a^{*i} > a^i \geq \ddot g^i$ and $\ddot a^{*j} < a^j \leq \ddot g^j$. Since $a$ is between $\ddot a^*$ and $\ddot g$ for both of the particles, we should investigate more to see how modifying $\ddot a^*$ changes the solution. \\
    
    The TV distance between $\ddot a^*$ and $a$ is $( |\ddot a^{*i} - a^i| + |\ddot a^{*j} - a^j| ) / 2 \geq ((1-\alpha^i) + \alpha^j) / 2$ for particles $i$ and $j$. The TV distance between $\ddot g$ and $a$ is $( |\ddot g^i - a^i| + |\ddot g^j - a^j| ) / 2 = (\alpha^i + (1-\alpha^j)) / 2$.  
    Since the particles are ordered with respect to their $\alpha^s$ values and $i$ and $j$ are placed to the opposite sides of the decision boundary, we know $\alpha^i \leq \alpha^j$. By using this fact, we see that changing $\ddot a^*$ (transferring $1$ multiplicity from $i$ to $j$), one obtains a closer solution to $\ddot g$, which is a better than or as good as the ``optimal'' solution. In either case, there is a contradiction. 
    \item \textbf{$i > \alpha$ and $j > \alpha$:} \\Both of the particles are placed at the right-hand side of the decision boundary. The TV reshuffling assigns the rounded down values as the multiplicities; $\ddot g^i = \floor{a^i}$ and $\ddot g^j = \floor{a^j}$. Combining these with the conditions in Equation \ref{sup_eq:tv_conditions}, we get $\ddot a^{*i} > a^i \geq \ddot g^i$ and $\ddot a^{*j} < \ddot g^j \leq a^j$. By changing $\ddot a^*$ (transferring $1$ multiplicity from $i$ to $j$), one obtains a closer solution to $\ddot g$, which is a better than or as good to the ``optimal'' solution. Either case, there is a contradiction. 
\end{enumerate}

Based on the ``$\ddot a^*$ is the optimal solution  closest to $\ddot g$'' assumption and the conditions in Equation \ref{sup_eq:tv_conditions}, we showed that; by modifying $\ddot a^*$, one can obtain a closer solution to $\ddot g$, which is a better than or as good to the ``optimal'' solution. This is a contradiction, therefore $\ddot g$ is an optimal solution.

\newpage
\section{Bootstrap Particle Filter with Likelihood}
The likelihood based reshuffling schemes ($\klp$ and $\tvp$) require changes in the standard BPF algorithm. The updated algorithm is shown in Algorithm \ref{sup_alg:bpf_likelihood}.

\begin{algorithm}[H]
   \caption{The Bootstrap Particle Filter with Likelihood}
   \label{sup_alg:bpf_likelihood}
\begin{algorithmic}
   \STATE {\bfseries Input:} $S$, $\{y_n\}_{n=1}^N$
   \STATE {\bfseries Output:} $q(x_{1:N})$
   \STATE initialize $x_1^s \sim\mu(x_1)$
   \STATE compute $\tilde w_1^s = g(y_1|x_1^s)$
   \STATE normalize $w_1^s = \tilde w_1^s / (\sum_s \tilde w_1^s)$
   \STATE compute $p(x^s_1, y_1) = g(y_1|x^s_1)\mu(x^s_1)$
   
   \FOR{$n=2,\ldots,N$}
   \IF{$\sum_s (w^s_{n-1})^{-2} < S / 2$}
   \STATE select ancestors $\{i^s_n\}_{s=1}^S= \mathcal{R}(\{p(x_{1:n}^s, y_{1:n})\}_{s=1}^S)$
   \STATE set $\tilde w^s_{n-1} = 1$, $\forall s$
   \ELSE 
   \STATE set $i^s_{n} = s$, $\forall s$
   \ENDIF
   \STATE propagate $x^s_n \sim f(x^s_n|x_{n-1}^{i^s_n})$
   \STATE compute $\tilde{w}_n^s = g(y_n|x_n^s) \tilde w_{n-1}^s$
   \STATE normalize $w_n^s = \tilde w_n^s / (\sum_s \tilde w_n^s)$
   \STATE compute $p(x_{1:n}^s,y_{1:n}) = g(y_n|x_n^s)f(x^s_n|x^{i^s_n}_{n-1})p(x_{1:n-1}^{i_n^s},y_{1:n-1})$
   \ENDFOR
   \STATE \textbf{return} $q(x_{1:N})$
\end{algorithmic}
\end{algorithm}

\section{Additional Results for SMC SV Experiments}
\begin{figure}[H]
\centering
\subfigure[$S=20$ vs. $S=500$.]{%
\label{sup_fig:l2_w_sampled_sv_20}%
\includegraphics[width=0.47\textwidth]{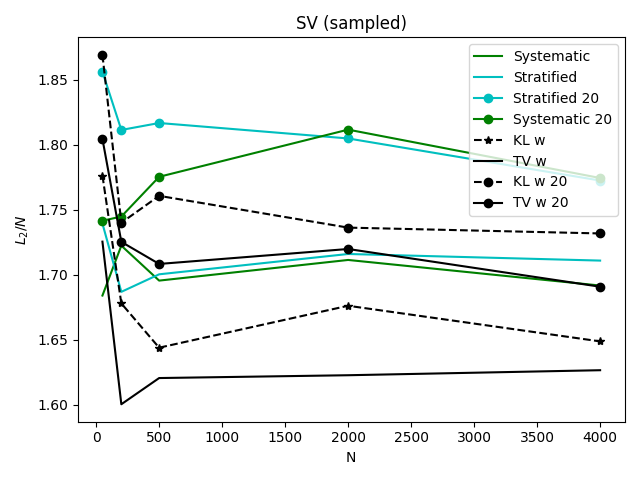}}%
\hskip 0.3cm
\subfigure[$S=50$ vs. $S=500$.]{%
\label{sup_fig:l2_w_sampled_sv_50}%
\includegraphics[width=0.47\textwidth]{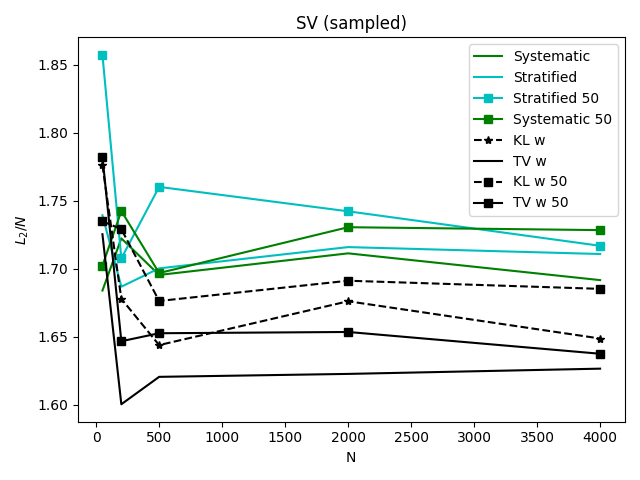}}%
\caption{\textbf{Sampled estimators}: Average $L_2$ loss per $N$ for weight based selection schemes. The losses are compared when using 20 (dash-circle) or 50 (dash-square) particles instead of 500.}
\label{sup_fig:SV_w_sampled_loss_vary_S}
\end{figure}

Here, we show the performance of the sampled estimators for $L_2$ loss. The left subplot of Figure \ref{sup_fig:SV_w_sampled_loss_vary_S} shows the performance comparison when the number of samples are $S=\{20, 500\}$. The right subplot of Figure \ref{sup_fig:SV_w_sampled_loss_vary_S} shows the performance comparison when the number of samples are $S=\{50, 500\}$. 

\section{Additional Results for SMC NL Experiments}

\begin{figure}[H]
\centering
\subfigure[$\theta=(1,1)$]{%
\label{sup_fig:nl_theta_11_L1}%
\includegraphics[width=0.25\textwidth]{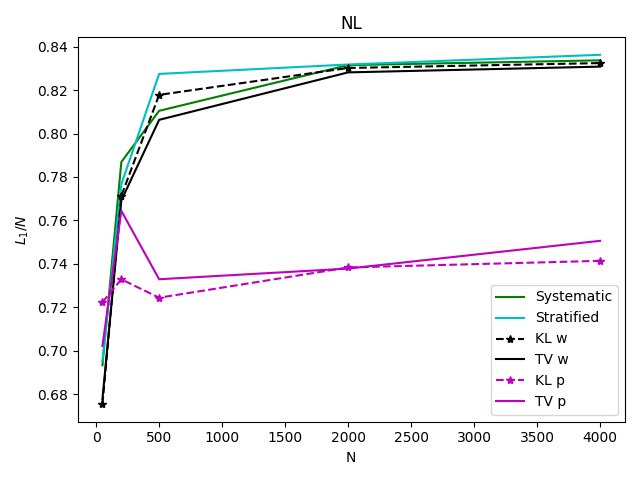}}%
\hskip 0.3cm
\subfigure[$\theta=(10,1)$]{%
\label{sup_fig:nl_theta_101_L1}%
\includegraphics[width=0.25\textwidth]{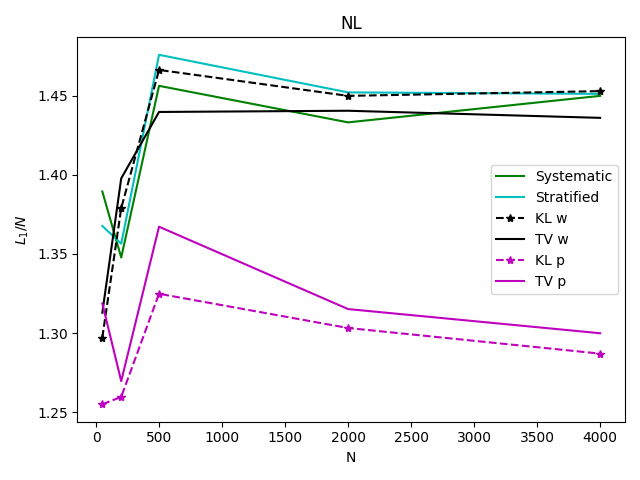}}%
\vskip\baselineskip
\subfigure[$\theta=(1, 10)$]{%
\label{sup_fig:nl_theta_110_l1}%
\includegraphics[width=0.25\textwidth]{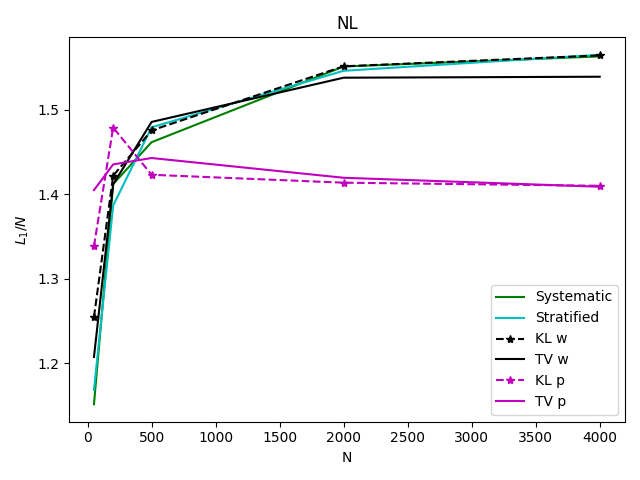}}%
\hskip 0.3cm
\subfigure[$\theta=(10, 10)$]{%
\label{sup_fig:nl_theta_1010_l1}%
\includegraphics[width=0.25\textwidth]{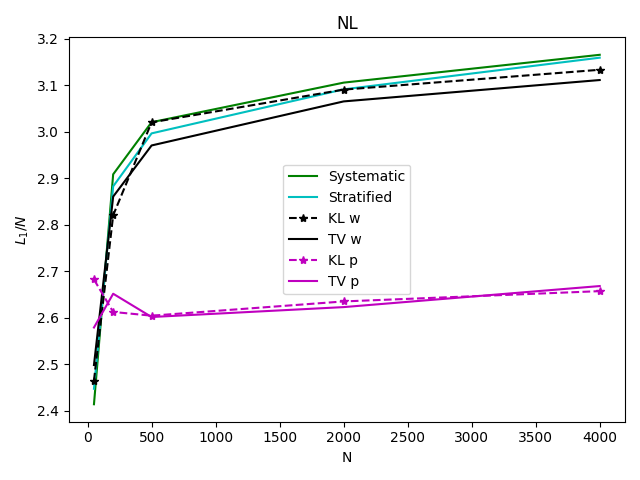}}%
\caption{\textbf{Bayesian estimators:} Average $L_1$ loss.}
\label{sup_fig:nl_l1_grid}
\end{figure}

\begin{figure}[H]
\centering
\subfigure[$\theta=(1,1)$]{%
\label{sup_fig:nl_theta_11_L2}%
\includegraphics[width=0.25\textwidth]{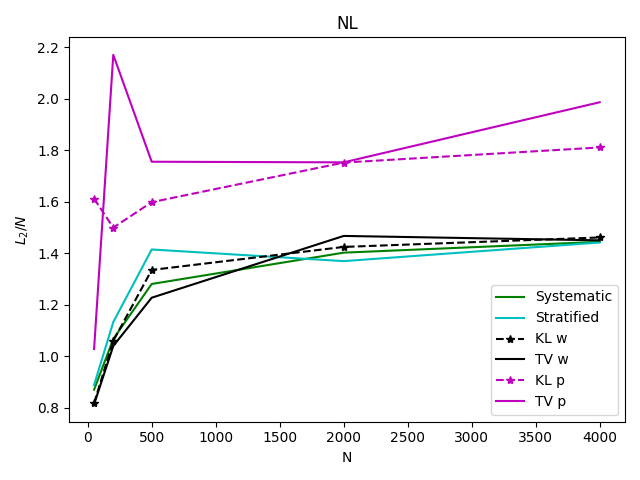}}%
\hskip 0.3cm
\subfigure[$\theta=(10,1)$]{%
\label{sup_fig:nl_theta_101_L2}%
\includegraphics[width=0.25\textwidth]{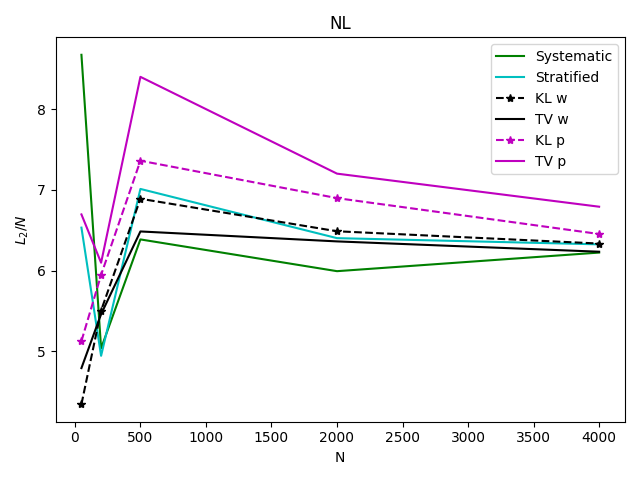}}%
\vskip\baselineskip
\subfigure[$\theta=(1, 10)$]{%
\label{sup_fig:nl_theta_110_l2}%
\includegraphics[width=0.25\textwidth]{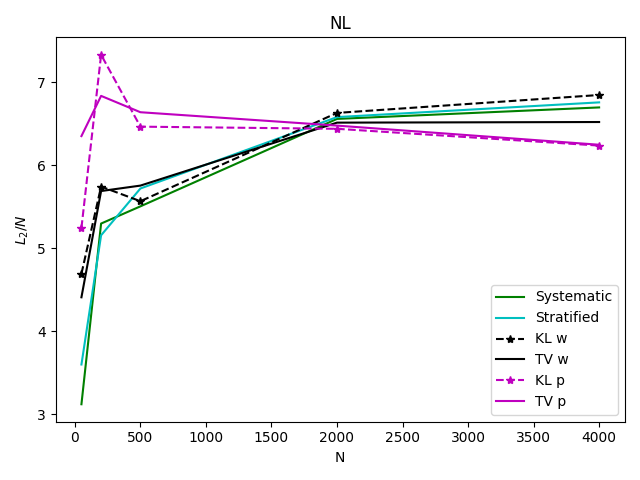}}%
\hskip 0.3cm
\subfigure[$\theta=(10, 10)$]{%
\label{sup_fig:nl_theta_1010_l2}%
\includegraphics[width=0.25\textwidth]{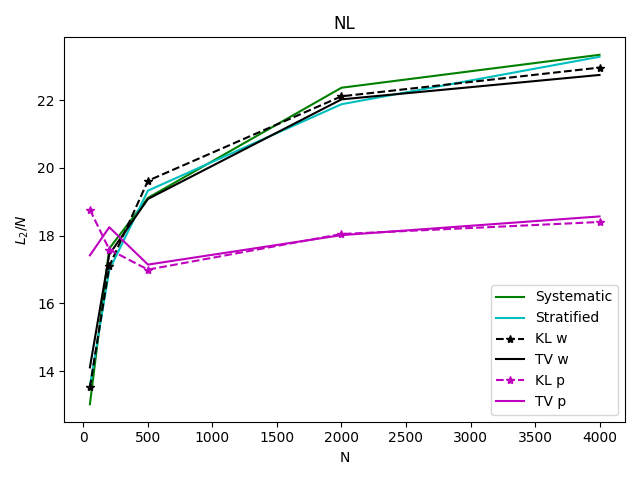}}%
\caption{\textbf{Bayesian estimators:} Average $L_2$ loss.}
\label{sup_fig:nl_l2_grid}
\end{figure}

Here, we present the performance of the Bayesian estimators for $L_1$ and $L_2$ losses (Figures \ref{sup_fig:nl_l1_grid} and  \ref{sup_fig:nl_l2_grid}, respectively) under different parameter settings. 

\newpage
\section{Conjugate Posterior Derivations for PG Experiments}
\label{sup_sec:posterior_derivations}
The derivations of the conjugate posteriors for $\sigma^2$ and $\beta$ are very similar, why we only give the derivations for $p(\sigma^2|x_{1:N})$.

First recall that
\begin{equation}
\label{sup_eq:app_ig_prior}
    \sigma^2 \sim \mathcal{IG}(\sigma^2|a, b) = \frac{b^a}{\Gamma(a)} \left(\sigma^2\right)^{(-a-1)} \exp\left\{\frac{-b}{\sigma^2}\right\},
\end{equation}
where $\Gamma$ is the Gamma function, and that $\sigma^2$ is independent of $y_{1:N}$ (as $\sigma^2$ is only involved in the transition distribution). This gives us 
\begin{equation}
\label{sup_eq:ig_post_sigma_app}
\begin{split}
    p(\sigma^2|x_{1:N}, y_{1:N})&= p(\sigma^2|x_{1:N}) \propto
    p(x_{1:N}|\sigma^2)p(\sigma^2)\\
    &= \mu(x_1)\prod_{n=2}^N f(x_{n}|x_{n-1})p(\sigma^2)
\end{split}
\end{equation}

Taking the logarithm on both sides of the above equation

\begin{equation}
\begin{split}
    \log p(\sigma^2|x_{1:N}) &= \log\mu(x_1) + \sum_{n=2}^N \log f(x_{n}|x_{n-1}) + \log p(\sigma^2)\\
    &=-\frac{1}{2}\log\frac{2\pi}{1-\phi^2} -\frac{1}{2}\log\sigma^2-\frac{1}{2}\frac{x_1^2}{\frac{\sigma^2}{1-\phi^2}}\\
    &\quad -\frac{N-1}{2}\log\sigma^2-\frac{N-1}{2}\log 2\pi -\frac{1}{2}\sum_{n=2}^N\frac{(x_n-\phi x_{n-1})^2}{\sigma^2}\\
    &\quad +a\log b-\log\Gamma(a) -(a+1)\log\sigma^2-\frac{b}{\sigma^2},
\end{split}
\end{equation}

followed by collecting all terms that are constant in $\sigma^2$ into the dummy variable $C_{\sigma^2}$

\begin{equation}
    \begin{split}
        \log p(\sigma^2|x_{1:N}) &=-\log\sigma^2\left(
        \frac{N}{2}+a+1
        \right)\\&\quad-\frac{1}{\sigma^2}\left[
        b+\frac{1}{2}x^2_1 (1-\phi^2) + \frac{1}{2}\sum_{n=2}^N(x_n-\phi x_{n-1})^2
        \right] + C_{\sigma^2},
    \end{split}
\end{equation}
and, finally, matching this expression to the general form of the $\mathcal{IG}$ distribution (see Equation \ref{sup_eq:app_ig_prior}), we arrive at the resulting expression
\begin{equation}
\label{eq:ig_post_sigma}
    p(\sigma^2|x_{1:N}) = \mathcal{IG}\left(\sigma^2|a + \frac{N}{2}, b + \frac{x_1^2(1 - \phi^2)}{2} + \frac{1}{2}\sum^{N}_{n=2} (x_n - \phi x_{n-1})^2\right).
\end{equation}

\newpage
\section{Additional Results for PG Synthetic Data Experiments}
\label{sup_sec:pg_synth}
\begin{figure}[H]
    \centering
    \includegraphics[width=10cm]{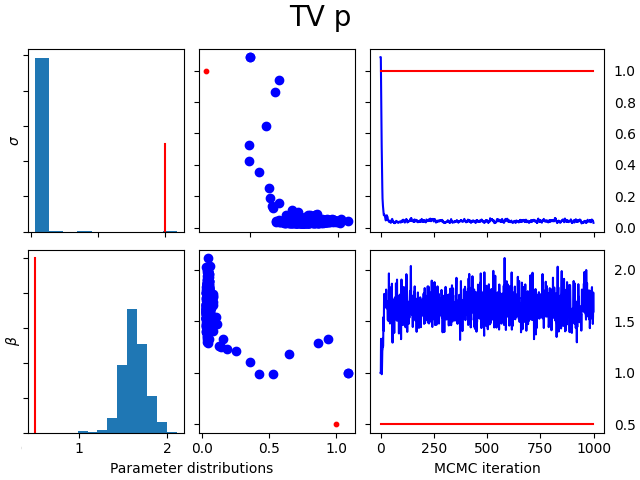}
    \caption{PG approximations of $p(\sigma^2|x_{1:N})$ and $p(\beta|y_{1:N})$ when using $\tvp$  ($N=100$, $S=10$, $M=1000$, no burn-in). Illustrated through histograms, scatter and trace plots.  Red markers indicated true parameter values.}
    \label{sup_fig:pmcmc_tv_p_hist}
\end{figure}

Here, we display a serious shortcoming of the likelihood based reshuffling schemes. As can be seen in Figure \ref{sup_fig:pmcmc_tv_p_hist}, the PG almost immediately converges to small variance values, far from the true variance, when using $\klp$ or $\tvp$.

In fact, by inspecting the SV model, especially the transition equation, the sensibility of the scheme becomes obvious. $\klp$ and $\tvp$ both reshuffle based on the transition probability, $p(x_n|x_{n-1}) = \mathcal{N}(x_n|\beta x_{n-1}, \sigma^2)$. 
Therefor, a smaller transition variance will encourage selection of particles located closest to the transition distribution's mean, in turn forcing the scale parameter in Equation \ref{eq:ig_post_sigma} to converge to $b$. This scale value places the mean and mode of the inverse gamma distribution far below 1 for any reasonable choice of $N$, which eventually, due to the zero mean in $\mu(x_1)$, causes $x_n\xrightarrow[]{m}0$ $\forall n\in[1,N]$, where $m$ denotes the MCMC iteration. Following this result, we will only focus on the weight based reshuffling schemes in this section (however, we continue discussing $\klp$ and $\tvp$ in the next chapter). Note, since the PG converged so rapidly to a local mode we only let $M=1000$ without burn-in.

\newpage
\section{Additional Results for PG Real Data Experiments}
\begin{figure}[H]
\centering
\subfigure[$\klw$, $S=20$]{%
\label{sup_fig:sp_2011_1}%
\includegraphics[width=0.23\textwidth]{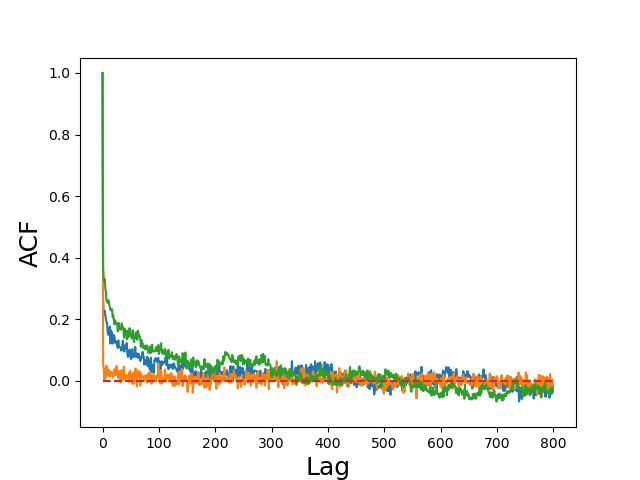}}%
\hskip 0.3cm
\subfigure[$\tvw$, $S=20$]{%
\label{sup_fig:sp_2011_2}%
\includegraphics[width=0.23\textwidth]{figs/SnP/ACFs2011/SnP_ACF_tv_w_N2011_S20_Iter30k.png}}%
\subfigure[Stratified, $S=20$]{%
\label{sup_fig:sp_2011_3}%
\includegraphics[width=0.23\textwidth]{figs/SnP/ACFs2011/SnP_ACF_stratified_N2011_S20_Iter30k.png}}%
\\
\subfigure[$\klw$, $S=50$]{%
\label{sup_fig:sp_2011_4}%
\includegraphics[width=0.23\textwidth]{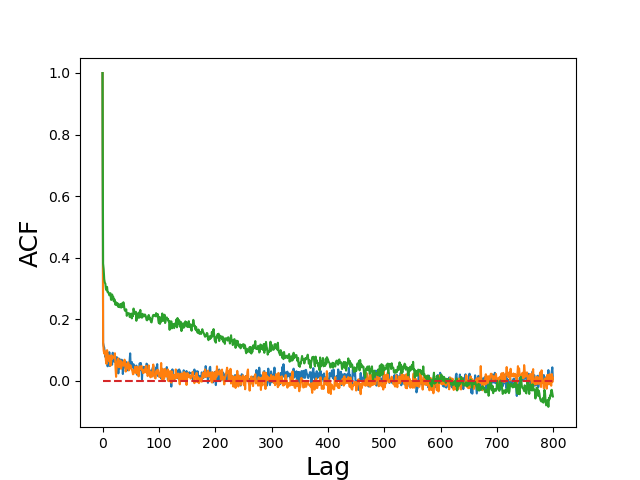}}%
\hskip 0.3cm
\subfigure[$\tvw$, $S=50$]{%
\label{sup_fig:sp_2011_5}%
\includegraphics[width=0.23\textwidth]{figs/SnP/ACFs2011/SnP_ACF_tv_w_N2011_S50_Iter30k.png}}%
\subfigure[Stratified, $S=50$]{%
\label{sup_fig:sp_2011_6}%
\includegraphics[width=0.23\textwidth]{figs/SnP/ACFs2011/SnP_ACF_stratified_N2011_S50_Iter30k.png}}%
\\
\subfigure[$\klw$, $S=100$]{%
\label{sup_fig:sp_2011_7}%
\includegraphics[width=0.23\textwidth]{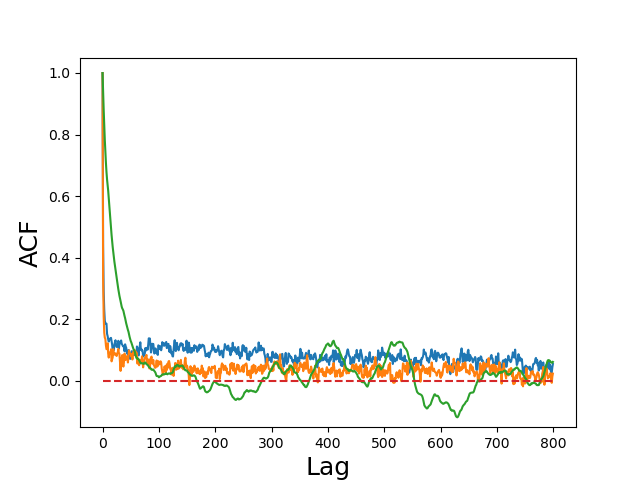}}%
\hskip 0.3cm
\subfigure[$\tvw$, $S=100$]{%
\label{sup_fig:sp_2011_8}%
\includegraphics[width=0.23\textwidth]{figs/SnP/ACFs2011/SnP_ACF_tv_w_N2011_S100_Iter30k.png}}%
\subfigure[Stratified, $S=100$]{%
\label{sup_fig:sp_2011_9}%
\includegraphics[width=0.23\textwidth]{figs/SnP/ACFs2011/SnP_ACF_stratified_N2011_S100_Iter30k.png}}%

\caption{ACF plots of $\klw$, $\tvw$ and stratified resampling for S\&P $N=2,011$ experiments. The columns are $\klw$ (left), $\tvw$ (middle) and stratified (right). The rows are the number of particles $S \in \{20, 50, 100\}$, respectively.}
\label{sup_fig:real_2011_acf}
\vskip -0.2in
\end{figure}

We present the ACF results of stratified, $\tvw$ and $\klw$ (not presented in the main text) in Figure \ref{sup_fig:real_2011_acf}. 

\section{Infrastructure Details}

All the runs are made on one of the 16 CPUs of an Intel Xeon Gold 6130 processor, which is part of a larger cluster. 

\newpage
\section{Particle Gibbs}

The particle Gibbs algorithm is shown in Algorithm \ref{sup_alg:pg}. At each iteration $m$, the algorithm i) samples the new parameters and ii) samples a trajectory w.r.t. the new parameters and the previous trajectory (see Algorithm \ref{sup_alg:pg_kernel}). The particle index sampled at the end of the kernel, $b$, is sampled from a Categorical distribution ($\mathcal{C}$) where the importance weights (or the particle likelihoods, depending on the version of the particle Gibbs) are the category probabilities. 

\begin{algorithm}[H]
   \caption{Particle Gibbs}
   \label{sup_alg:pg}
\begin{algorithmic}
\STATE {\bfseries Input:} $S$, $\{y_n\}_{n=1}^N$
   \STATE {\bfseries Output:} $\theta^M, x_{1:N}^M$
   \STATE initialize $x_{1:N}^1$ and $\theta^1$

   \FOR{$m=2,\ldots,M$}
   \STATE sample parameter $\theta^m \sim p(\theta|y_{1:N}, x_{1:N}^{m-1})$
   \STATE sample trajectory $x_{1:N}^m \sim \mathcal{K}(S, y_{1:N}, \theta^m,  x_{1:N}^{m-1})$
   \ENDFOR
   \STATE \textbf{return} $\theta^M, x_{1:N}^M$
\end{algorithmic}
\end{algorithm}

\begin{algorithm}[H]
   \caption{Particle Gibbs Kernel, $\mathcal{K}$}
   \label{sup_alg:pg_kernel}
\begin{algorithmic}
   \STATE {\bfseries Input:} $S$, $\{y_n\}_{n=1}^N$, $\theta$, $\mathbf{x}_{1:N}$
   \STATE {\bfseries Output:} $x^b_{1:N}$
   \STATE initialize $x_1^s \sim\mu(x_1)$
   \STATE set $x_1^S = \mathbf{x}_1$
   \STATE compute $w_1^s = 1 / S$, $\forall s$
   
   \FOR{$n=2,\ldots,N$}
   \IF{$\sum_s (w^s_{n-1})^{-2} < S / 2$}
   \STATE select ancestors $\{i^s_n\}_{s=1}^{S}= \mathcal{R}(\{w^s_n\}_{s=1}^{S})$
   \STATE set $i^S_n = S$
   \STATE set $\tilde w^s_{n-1} = 1$, $\forall s$
   \ELSE 
   \STATE set $i^s_{n} = s$, $\forall s$
   \ENDIF
   \STATE propagate $x^s_n \sim f(x^s_n|x_{n-1}^{i^s_n}, \theta)$
   \STATE set $x^S_n = \mathbf{x}_n$
   \STATE compute $\tilde{w}_n^s = g(y_n|x_n^s, \theta) \tilde w_{n-1}^s$ 
   \STATE normalize $w_n^s = \tilde w_n^s / (\sum_s \tilde w_n^s)$
   \ENDFOR
   \STATE draw $b = \mathcal{C}(\{w^s_n\}_{s=1}^S)$
   \STATE \textbf{return} $x^b_{1:N}$
\end{algorithmic}
\end{algorithm} 

For the likelihood based particle Gibbs, one should make the necessary updates, just like the BPF case. 

\newpage
\section{Loss Function and Their Estimators}

Table \ref{sup_tab:loss} shows the loss functions used in this study and their Bayesian estimators. 

\begin{table}[H]
\caption{Loss functions and their estimators}
\label{sup_tab:loss}
\vskip 0.15in
\begin{center}
\begin{small}
\begin{sc}
\begin{tabular}{cccc}
\toprule
Loss Function Name & Loss Function & Estimator Name & Estimator Formula \\
\midrule
0-1 Loss & $L_{0-1}(x_n, \hat x_n)$ & Maximum a posteriori (MAP) & $\hat x_n^\text{MAP} = \arg \max p(x_{n}|y_{1:N})$ \\
Absolute Loss & $L_1(x_n, \hat x_n)$ & Minimum mean absolute error (MMAE) & $\hat x_n^\text{MMAE} = med\left[p(x_{n}|y_{1:N})\right]$ \\
Quadratic Loss & $L_2(x_n, \hat x_n)$ & Minimum mean squared error (MMSE) & $\hat x_n^\text{MMSE} = \sum_{s}w^s x_n^s$ \\
\bottomrule
\end{tabular}
\end{sc}
\end{small}
\end{center}
\vskip -0.1in
\end{table}

where $med$ is the median. 

The binary loss is 
\begin{equation*}
    L_{0-1}(x_n, \hat x_n) = 
    \begin{cases}
    0& : \quad |x_n-\hat x_n| \leq \sigma / 2 \\
    1&: \quad |x_n-\hat x_n| > \sigma / 2 \\
    \end{cases}.
\end{equation*}

The absolute loss is 
\begin{equation*}
    L_1(x_n, \hat x_n) = |x_n - \hat x_n|.
\end{equation*}

The quadratic loss is 
\begin{equation*}
    L_2(x_n, \hat x_n) = (x_n - \hat x_n)^2.
\end{equation*}

\end{document}